\documentclass{article}
\usepackage{subcaption}

\title{Mechanistic Design and Scaling of Hybrid Architectures}
\author{
    Michael Poli\footnote{Equal contribution.}$^{*,1,7}$, Armin W Thomas$^{*,2,7}$, Eric Nguyen$^{*,2}$, \\ Pragaash Ponnusamy$^1$, Björn Deiseroth$^3$, Kristian Kersting$^3$, Taiji Suzuki$^4$, \\ Brian Hie$^{2,5}$, Stefano Ermon$^{2,6}$, Christopher R\'e$^2$, Ce Zhang$^1$, Stefano Massaroli$^{4, 7}$
 }
\date{%
    \small $^1$Together AI,~$^2$Stanford University,~$^3$Hessian AI,~$^4$RIKEN,~$^5$Arc Institute,~$^6$CZ Biohub,~$^7$Liquid AI
}
\usepackage[T1]{fontenc}  
\usepackage[bookmarks=true, 
            colorlinks=true,
            linkcolor=bluegray,
            citecolor=junglegreen]{hyperref}
\usepackage[backref=true, 
            sorting=none]{biblatex}
\usepackage[a4paper, margin=.9in]{geometry}
\usepackage{wrapfig}
\usepackage{booktabs}  
\usepackage[bottom]{footmisc}
\usepackage{enumitem} 

\usepackage{chngcntr}
\counterwithin{figure}{section} 
\counterwithin{table}{section}

\usepackage{graphicx}
\usepackage{tikz}
\usepackage{tikz-cd}
\usepackage{hf-tikz}
\usepackage{pgfplots} 
\pgfplotsset{compat=1.17} 
\pgfplotsset{
        table/search path={figures/drawings},
    }
\usetikzlibrary{fadings}
\usetikzlibrary{shapes, arrows, fit, backgrounds, arrows.meta}

\usetikzlibrary{matrix}
\usetikzlibrary{shadows.blur}
\usetikzlibrary{patterns, tikzmark}
\usetikzlibrary{decorations.pathreplacing, calc, decorations.markings,}
\usetikzlibrary{positioning}

\usepgfplotslibrary{groupplots}
\usepgfplotslibrary{patchplots}
\definecolor{bluegray}{rgb}{0.4, 0.6, 0.8}
\definecolor{bluebell}{rgb}{0.64, 0.64, 0.82}
\definecolor{etonblue}{rgb}{0.59, 0.78, 0.64}
\definecolor{junglegreen}{rgb}{0.16, 0.67, 0.53}
\definecolor{bg}{gray}{0.97}
\definecolor{olive}{rgb}{0.6, 0.6, 0.2}
\definecolor{sand}{rgb}{0.8666666666666667, 0.8, 0.4666666666666667}
\definecolor{wine}{rgb}{0.5333333333333333, 0.13333333333333333, 0.3333333333333333}
\definecolor{deblue}{RGB}{11,132,147}
\definecolor{ocra}{RGB}{204, 119, 34}

\pgfplotsset{%
            mesh line legend/.style={legend image code/.code=\meshlinelegend#1},%
}
\makeatletter
\long\def\meshlinelegend#1{%
    \scope[%
        #1,
        /pgfplots/mesh/rows=1,
        /pgfplots/mesh/cols=4,
        /pgfplots/mesh/num points=,
        /tikz/x={(0.44237cm,0cm)}, 
        /tikz/y={(0cm,0.23932cm)},
        /tikz/z={(0.0cm,0cm)},
        scale=0.4,
    ]
    \let\pgfplots@metamax=\pgfutil@empty
    \pgfplots@curplot@threedimtrue

    \pgfplotsplothandlermesh
    \pgfplotstreamstart

    \def\simplecoordinate(##1,##2,##3){%
        \pgfmathparse{1000*(##3)}%
        \pgfmathfloatparsenumber\pgfmathresult
        \let\pgfplots@current@point@meta=\pgfmathresult
        \pgfplotstreampoint{\pgfqpointxyz@orig{##1}{##2}{##3}}%
    }%

    \pgfplotsforeachungrouped \x in {0,...,\pgfkeysvalueof{/pgfplots/samples}}{
        \pgfmathsetmacro\y{\x/\pgfkeysvalueof{/pgfplots/samples}}
        \pgfmathsetmacro\x{\x/\pgfkeysvalueof{/pgfplots/samples}*3}
        \simplecoordinate(\x,0,\y)
    }

    \pgfplotstreamend
    \pgfusepath{stroke}
    \endscope
}%
\makeatother

\usepackage{lettrine}
\usepackage{Typocaps}

\LettrineTextFont{\itshape}
\usepackage{amsthm}

\usepackage{minitoc}

\newcommand{\chapref}[1]{\hyperref[#1]{Chapter \ref{#1}}}
\newcommand{\secref}[1]{\hyperref[#1]{Section \ref{#1}}}

\usepackage{marginnote}

\usepackage[many]{tcolorbox}
\tcbuselibrary{breakable,xparse,skins}

\usepackage{newunicodechar}
\newunicodechar{λ}{{$\mathtt\lambda$}}
\newunicodechar{μ}{{$\mathtt\mu$}}

\usepackage{xparse}
\DeclareTColorBox{emphbox}{O{black}O{0cm}}{
    enhanced jigsaw,
    breakable,
    outer arc=0pt,
    arc=0pt,
    colback=white,
    rightrule=0pt,
    toprule=0pt,
    top=0pt,
    right=0pt,
    bottom=0pt,
    bottomrule=0pt,
    colframe=#1,
    enlarge left by=#2,
    width=\linewidth-#2,
}

\DeclareTColorBox{emphbox}{O{black}O{0cm}}{
    empty,
    breakable=true,
    outer arc=0pt,
    arc=0pt,
    rightrule=0pt,
    leftrule=2pt,
    borderline west={2pt}{0pt}{#1},
    toprule=0pt,
    top=0pt,
    right=-3pt,
    bottom=0pt,
    bottomrule=0pt,
    colframe=#1,
    enlarge left by=#2,
    width=\linewidth-#2,
}
\usepackage{listings}
\makeatletter
\newcommand\BeraMonottfamily{%
  \def\fvm@Scale{0.85}
  \fontfamily{fvm}\selectfont
}
\makeatother

\usepackage[normalem]{ulem}

 \usepackage[draft,inline,nomargin,index]{fixme}
\fxsetup{theme=color,mode=multiuser}
\FXRegisterAuthor{sm}{asm}{\color{blue!70}SM}
\usepackage{amsmath, amssymb, amsfonts, mathtools}
\usepackage{physics}
\usepackage{cancel}
\usepackage{thmtools, thm-restate}
\usepackage{accents}
\usepackage{bm}

\makeatletter
\newcommand{\ostar}{\mathbin{\mathpalette\make@circled *}}
\newcommand{\make@circled}[2]{%
  \ooalign{$\m@th#1\smallbigcirc{#1}$\cr\hidewidth$\m@th#1#2$\hidewidth\cr}%
}
\newcommand{\smallbigcirc}[1]{%
  \vcenter{\hbox{\scalebox{0.77778}{$\m@th#1\bigcirc$}}}%
}
\makeatother

\newcommand{\Wsf}{\mathsf{W}}
\newcommand{\Wsfr}{{\color{blue!70}\mathsf{W}}}

\newcommand{\x}{\times}

\usepackage{pict2e, picture}
\makeatletter
\DeclareRobustCommand{\Arrow}[1][]{%
\check@mathfonts
\if\relax\detokenize{#1}\relax
\settowidth{\dimen@}{$\m@th\rightarrow$}%
\else
\setlength{\dimen@}{#1}%
\fi
\sbox\z@{\usefont{U}{lasy}{m}{n}\symbol{41}}%
\begin{picture}(\dimen@,\ht\z@)
\roundcap
\put(\dimexpr\dimen@-.7\wd\z@,0){\usebox\z@}
\put(0,\fontdimen22\textfont2){\line(1,0){\dimen@}}
\end{picture}%
}
\makeatother

\newcommand{\cL}{\mathcal{L}}
\newcommand{\cM}{\mathcal{M}}

\newcommand{\R}{\mathbb{R}}

\DeclareMathAlphabet{\nummathbb}{U}{BOONDOX-ds}{m}{n}

\makeatletter
\DeclareRobustCommand\widecheck[1]{{\mathpalette\@widecheck{#1}}}
\def\@widecheck#1#2{%
    \setbox\z@\hbox{\m@th$#1#2$}%
    \setbox\tw@\hbox{\m@th$#1%
       \widehat{%
          \vrule\@width\z@\@height\ht\z@
          \vrule\@height\z@\@width\wd\z@}$}%
    \dp\tw@-\ht\z@
    \@tempdima\ht\z@ \advance\@tempdima2\ht\tw@ \divide\@tempdima\thr@@
    \setbox\tw@\hbox{%
       \raise\@tempdima\hbox{\scalebox{1}[-1]{\lower\@tempdima\box
\tw@}}}%
    {\ooalign{\box\tw@ \cr \box\z@}}}
\makeatother

\begin{document}
\numberwithin{equation}{section}
\maketitle

\begin{abstract} 
The development of deep learning architectures is a resource-demanding process, due to a vast design space, long prototyping times, and high compute costs associated with at-scale model training and evaluation. We set out to simplify this process by grounding it in an end-to-end \textit{mechanistic architecture design} ({\tt MAD}) pipeline, encompassing small-scale capability unit tests predictive of scaling laws. Through a suite of synthetic token manipulation tasks such as compression and recall, designed to probe capabilities, we identify and test new hybrid architectures constructed from a variety of computational primitives. We experimentally validate the resulting architectures via an extensive compute-optimal and a new state-optimal scaling law analysis, training over $500$ language models between $70\text{M}$ to $7\text{B}$ parameters. Surprisingly, we find {\tt MAD} synthetics to correlate with compute-optimal perplexity, enabling accurate evaluation of new architectures via isolated proxy tasks. The new architectures found via {\tt MAD}, based on simple ideas such as hybridization and sparsity, outperform state-of-the-art Transformer, convolutional, and recurrent architectures (Transformer{\tt ++}, Hyena, Mamba) in scaling, both at compute-optimal budgets and in overtrained regimes. Overall, these results provide evidence that performance on curated synthetic tasks can be predictive of scaling laws, and that an optimal architecture should leverage specialized layers via a hybrid topology. 
\end{abstract}
\section{Introduction}

Alongside data quality, the effectiveness of large-scale training is determined by the quality of a model architecture \cite{kaplan2020scaling,hoffmann2022training}, which is defined by the set and arrangement of the computational primitives used to form layers and functional blocks, as well as their parametrization.

Due to the combinatorial explosion of possible architecture designs and a lack of reliable prototyping pipelines -- despite progress on automated neural architecture search methods \cite{white2023neural} -- architectural improvements are obtained through an opaque development process guided by heuristics and individual experience, rather than systematic procedures.
Further adding to this issue are the large costs and long iteration times associated with training and testing new architectures, underscoring the need for principled and nimble design pipelines. 

In spite of the wealth of possible architecture designs, the majority of models rely on variations of the same uniform Transformer recipe, based on a regular interleaving of memory-based mixers (self-attention layers) with \textit{memoryless} mixers (shallow FFNs) \cite{touvron2023llama,jiang2023mistral}.
This particular combination of computational primitives -- originating from the first Transformer design \cite{vaswani2017attention} -- is known to improve quality, with empirical arguments supporting the notion that these primitives specialize in different sequence modeling sub-tasks e.g., in-context versus factual recall \cite{geva2023dissecting}.
Beyond the Transformer architecture are a class of emerging computational primitives inspired by signal processing, based on gated convolutions and recurrences \cite{katharopoulos2020transformers,peng2023rwkv,poli2023hyena,nguyen2023hyenadna,gu2023mamba,yang2023gated}, promising improved quality, cheaper scaling to long sequence length, and efficient inference.
These new primitives expand the architecture design space, offering new opportunities to extend capabilities and specializations of models.

In this work, we set out to explore key questions arising from these observations:

\begin{enumerate}
    \item \textit{Can the architecture design process be streamlined through a set of simple pretext token manipulation tasks, providing quick and cheap performance estimates predictive of scaling laws?}
    \item \textit{Is it possible to bring together the ``best of all worlds'' by arranging different computational primitives into hybrid architecures, leveraging their respective specialized capabilities?}
\end{enumerate}

In an attempt to provide answers to these questions, we make the following core contributions:
\begin{figure}[t]
    \centering
    \includegraphics[width=0.9\linewidth]{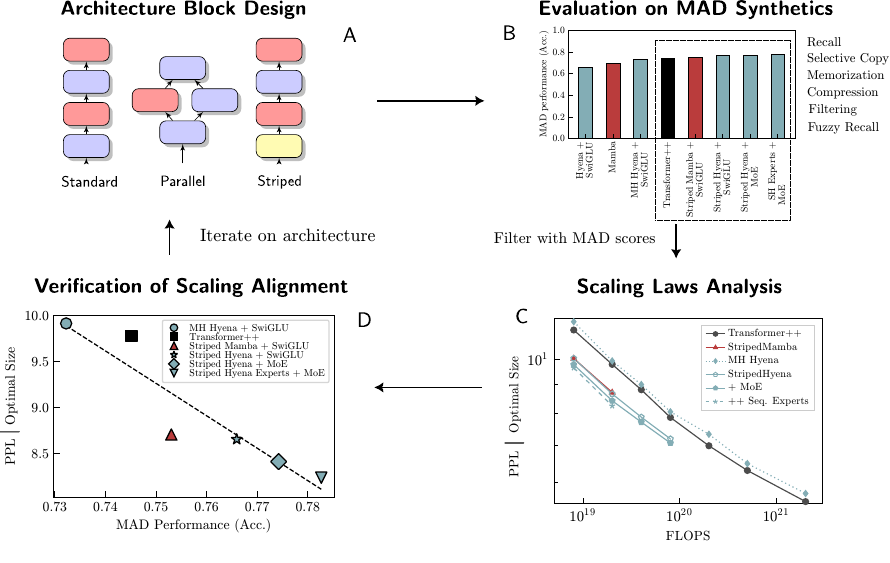}
    \vspace{-3mm}
    \caption{
    \textit{Mechanistic architecture design} ({\tt MAD}) is a framework to enable fast iterative improvement of architectures, including emerging approaches based on recurrences and convolutions. \textbf{[A]:} Design architectures via selection of computational primitives and topology. \textbf{[B]:} {\tt MAD} involves an evaluation of architecture designs at small scale on a set of token manipulation synthetic tasks, curated to unit test a variety of model capabilities. The experimental setup promotes direct comparison via normalization of total state dimension for recurrent models. \textbf{[C]:} Validate scaling laws of top-performing models on {\tt MAD} synthetics in compute-optimal and overtrained regimes. Results in B used to reduce the number of candidate architectures. \textbf{[D]:} Verify alignment of scaling properties and ({\tt MAD}) results for each architecture e.g., correlation of compute-optimal scaling perplexity and aggregate ({\tt MAD}) score (in the figure, compute-optimal perplexity at 2e19 FLOP budget is shown). If the scores between target quantity and ({\tt MAD}) synthetics are correlated, iterate on a single target architecture.}
    \vspace{-0.2in}
    \label{fig:catchy}
\end{figure}
\paragraph{Mechanistic architecture design}
We introduce a methodology for the fast prototyping and testing of new architectures, \textit{mechanistic architecture design} ({\tt MAD}). {\tt MAD} is a collection of synthetic tasks -- such as recall, memorization, and compression -- curated to serve as isolated unit tests for key capabilities of an architecture, requiring only minutes of training time. In particular, {\tt MAD} tasks are inspired by progress on understanding the inner workings of Transformers and other sequence models via in-context learning, recall, and other sequence manipulation tasks \cite{olsson2022context,fu2022hungry,bhattamishra2023understanding,arora2023zoology,akyurek2024context}. We apply {\tt MAD} to test architectures built with representative computational primitives such as gated convolutions \cite{poli2023hyena}, gated input-varying linear recurrences \cite{gu2023mamba,yang2023gated}, and other operators e.g., \textit{mixture of experts} (MoEs) \cite{shazeer2017outrageously}, as well as novel ones. With {\tt MAD}, we are able to filter for promising architecture candidates (Fig, \ref{fig:catchy}, \textbf{[A,B]}). By identifying which individual tasks computational primitives excel at, we find and validate several ways to improve designs, such as \textit{striping} i.e., sequentially interleaving blocks composed of different computational primitives with a specified interconnection topology, resulting in hybrid architectures \cite{ma2022mega, fu2022hungry,fathi2023block}.
\paragraph{Scaling laws of emerging architectures}
To investigate the link between {\tt MAD} synthetics and real-world scaling, we execute the \textbf{largest scaling law analysis on emerging architectures to date}, training over $500$ language models between $70$ million and $7$ billion parameters with different architectures. Our protocol builds and expands on compute-optimal scaling laws for LSTMs and Transformers \cite{kaplan2020scaling,stanic2023languini,hoffmann2022training}. Our findings show that hybrid architectures improve on all scaling measures, resulting in lower pretraining losses at different \textit{floating point operation} (FLOP) compute-budgets at the compute-optimal frontier\footnote{Found via the optimal allocation of compute to tokens and model size.}. We also verify new architectures to be more robust to large pretraining runs outside the efficient frontier e.g., smaller models trained for significantly more tokens, which make up a majority of training settings in practice due to inference cost considerations \cite{sardana2023beyond}. 
\paragraph{Hybridization insights at scale}
Building on our scaling law analysis, we investigate hybridization schedules and model topology. Our findings uncover optimal hybridization ratios for attention~\cite{vaswani2017attention}, Hyena~\cite{poli2023hyena}, and Mamba~\cite{gu2023mamba} mixtures, as well as the respective placement of these layers in an architecture. 
\paragraph{State-optimal scaling laws}
The size of the \textit{state} -- the analog of \textit{kv-caches} in standard Transformers \cite{massaroli2023laughing} -- of emerging convolutional and recurrent primitives \cite{poli2023hyena,gu2023mamba} plays a central role in {\tt MAD} and our scaling analysis, as it determines inference efficiency, memory cost, and provably has a direct effect on recall capabilities \cite{arora2023zoology}. We introduce a \textit{state-optimal scaling} analysis, with the objective of estimating how perplexity scales with the state dimension of different model architectures. We find hybrid architectures to balance the trade-off between compute requirements, state dimension, and perplexity.

\paragraph{New state-of-the-art architectures}
Leveraging {\tt MAD} and new computational primitives, derived from the insights developed in this work, we design new state-of-the-art hybrid architectures, outperforming the best Transformer, convolutional, and recurrent baselines (Transformer{\tt ++} \cite{touvron2023llama}, Hyena, Mamba) with a reduction of up to $20\%$ in perplexity for the same compute budget.

\paragraph{Correlation between synthetics and scaling performance}
Finally, we provide the first evidence that a curated selection of {\tt MAD} synthetic tasks can be used to reliably predict scaling law performance, paving the way to faster, automated architecture design. In particular, {\tt MAD} accuracy is rank-correlated with compute-optimal perplexity at scale (Fig. \ref{fig:catchy}, [D]), with particularly strong correlation for models in the same architecture class (Fig \ref{fig:mad-to-scale}). 

\section{Background: Architecture Design}\label{sec:background}
Architecture design refers to the selection and optimization of (a) computational primitives and their composition into layers and blocks, and (b) topology i.e., the interconnection and placement of individual blocks in an architecture.

In the following, we define the bounds of the architecture design search space explored in this work. In particular, we provide details on the emerging class of implicit subquadratic models, since their properties drive the design of the synthetic task and evaluation pipeline in {\tt MAD}, and motivate the introduction of a state-optimal scaling law analysis.

\subsection{Computational primitives}
Architectures are compositions of linear and nonlinear functions with {\color{blue!70}learnable parameters}. Common choices for the former are parametric dense or structured layers ${\mathsf L}: \R^T \rightarrow \R^T$, $y = \mathsf{L}(u)$. As an example,
\[
\begin{aligned}
&{\sf dense}  \quad &y_t &= \sum_{t'=1}^{T} {\color{blue!70}\Wsfr_{t t'}}u_{t'}, \quad &\Wsfr \in \R^{T \times T} \\
&{\sf (causal)~ conv.} \quad &y_t &= \sum_{t'=1}^{t} {\color{blue!70}\Wsfr_{t - t'}}u_{t'}, \quad &\Wsfr \in \R^{T}.\\
\end{aligned}
\]
It is often useful to differentiate between explicitly and implicitly parametrized layers, depending on whether the entries $\Wsf_{tt'}$ are the learnable parameters of the layer or are themself parametric functions of positional encodings or of the input, i.e. $(t,t',u)\mapsto \Wsf_{tt'}(u)$ \cite{poli2023hyena}. Implicit parametrizations disentangle the number of model parameters and dimensionality $T$ of the inputs. Further, they can be leveraged to create complex dependencies on the inputs in the entries of $\Wsf(u)$ such as in self-attention, $\Wsf_{tt'}(u) = \sigma(\langle Qu_t,Ku_{t'}\rangle)$. This ensures the layer can be applied to inputs with large $T$ without a prohibitive parameter and memory cost. We often refer to the implicit parametrization for an implicit layer as its \textit{featurization path}.

\paragraph{On nonlinearities in architecture design}
Linear primitives are typically interconnected via nonlinearities and residuals. Common nonlinearities are applied elementwise or to some specific dimension (e.g., the softmax used in attention). \cite{lin2017structured,vaswani2017attention}. Another commonly employed nonlinearity is gating, resulting in a polynomial function of the input. While other lines of work investigate choice and placement of nonlinearities in a layer to optimize quality, efficiency, or to minimize the emergence of activation outliers \cite{so2021primer}, these quality improvements are smaller compared to other layer and topology changes\footnote{Many tweaks to activation choice, placement and presence of biases are carried out to improve numerical stability and reduce the presence of large outliers in activations, rather than improve scaling performance.} and are thus outside the scope of this work.

\paragraph{Implicit primitives}
Implicitly parametrized computational primitives are the backbone of most model architectures of practical interest. An important class of implicit layers can be described starting from so-called \textit{linear attention} \cite{katharopoulos2020transformers,schlag2021linear,hua2022transformer}\footnote{We use $t$ for consistency, although in practice these layers can be applied to both "sequence" dimension, as well as "width" dimension.}, in its simplest (single-channel, unnormalized\footnote{For simplicity we detail unnormalized layers, as normalization simply redefines the operator as the ratio of two recurrences.}) form 
\begin{equation}\label{eq:linatt}
    \begin{aligned}   
    &{\sf recurrence} &x_{t+1} &= x_{t} + k_{t}(u) v_{t}(u) \\ 
    &{\sf readout} &y_{t} &= q_{t}(u) x_{t}
    \end{aligned}
\end{equation}
where $q, k, v: \R^T \rightarrow \R^T$ are the featurization path of the layer. Linear attention is a linear \textit{recurrent neural network} (RNN) or \textit{state-space model} (SSM) with constant identity state-to-state dynamics, and implicitly-parametrized input-to-state and state-to-output mappings. Linear attention can be evaluated in parallel during training or inference prefilling using its parallel form $y_t = q_t \sum_{t'=1}^t k_{t'} v_{t'}$, without materializing the state $x$. Notably, the class of subquadratic implicit models \cite{poli2023hyena,gu2023mamba,yang2023gated} emerges as generalizations of \eqref{eq:linatt} with a few key differences.

\subsection{State, cache, and memory}\label{taxonomy}
In autoregressive tasks, such as text generation, recurrent models enable lower latency and constant memory generation, since the fixed state $x_t$  replaces the cache required in other generic nonlinear blocks such as attention e.g., the \textit{$kv$-cache}. Indeed, $kv$-caches can be seen as a state of dynamic size, by reformulating attention as a recurrence with state size $T$, see \cite{massaroli2023laughing}. For this reason, we use fixed states and dynamic states to refer to states and $kv$-caches in hybrid architectures.

\paragraph{Nonparametric state expansion tricks}

The size of the state and its utilization play a central role in the taxonomy, analysis, and design of efficient architectures. State size, as well as the parametrization of a block, determine memorization and recall capabilities of a layer, as well as inference efficiency. For this reason, different approaches have been developed to expand the state dimension without prohibitive parameter cost. The main ones are the \textit{outer-product head trick}: 

\[
\begin{aligned}   
    x_{t+1} &= x_{t} + (k_{t} \otimes I_M) v_t,\quad &k_t, v_t, q_t &\in \R^M \\ 
    y_{t} &= (I_M \otimes q_t) x_{t},\quad &x_t &\in \R^{M^2}.
\end{aligned}
\]

Note that we have used a vectorized notation instead of the commonly employed matrix notation for models using the state expansion trick. This configuration linearly increases the state size from a head dimension $M$ to a total of $M^2$, and is employed in most linear attention variants \cite{katharopoulos2020transformers}, Hyena and RWKV variants \cite{massaroli2023laughing,peng2023rwkv} as well as GLA \cite{yang2023gated}. 

The second method to expand the total number of states per layer is achieved via the \textit{multi single-input single-output} (mSISO) layer configuration, which is equivalent to applying multiple independent recurrences with $M$ states in parallel.

Given the importance of the total state dimension in determining the capacity of a layer, we find \textbf{model comparisons in an \textit{iso-state} setting} -- normalizing for the total number of states regardless of the specifics of the layer -- to be required to ensure architecture improvements measured on smaller scale synthetic tasks can transfer to pretraining results at scale.

\paragraph{Manipulating the state}
Beyond state expansion techniques, efficient layers can be taxonomized based on their parametrization of state-to-state dynamics and their implicit parameters. For example, an input-varying layer introduces additional featurization path to extend input-variance to state-to-state transitions e.g., $x_{t+1} = {\color{blue!70}g_t(u)} x_{t} + k_{t}(u) v_t(u)$. We choose three state-of-the-art approaches spanning different possible combinations:
{}
\begin{center}
    \begin{tabular}{r|l|l}
     Hyena \cite{poli2023hyena} & \textit{weakly} input-varying\footnotemark& mSISO\\
     Multi-Head Hyena \cite{massaroli2023laughing} 
     & \textit{weakly} input-varying & mSISO with heads\\
     \midrule
     Gated Linear Attention \cite{yang2023gated} & input-varying & heads\\
     \midrule
     Mamba \cite{gu2023mamba} & input-varying & mSISO and weight sharing\footnotemark
    \end{tabular}
\end{center}
\footnotetext{Only the input-to-state and output-to-state maps are input-varying.}
\footnotetext{Input-to-state and state-to-output maps are shared across channels.}


The layers also vary slightly in their featurization paths e.g., GLA uses a low-rank elementwise implicit state-to-state transition, whereas Mamba uses a different low-rank parametrization and weight-tying.

\subsection{Topology}
Beyond the specifics of the layer itself, designing architectures involves arranging these computational primitives into blocks, interconnected with a particular topology, for example, sequential, parallel, or hybrid (as illustrated in Fig.~\ref{fig:catchy}). In this work, we explore sequential striped topologies i.e., where different computational primitives are applied sequentially, as well as sparse parallel topologies i.e., mixture of experts.

\section{Mechanistic Architecture Design}
\label{mechdes}

In the ideal case, we would have access to an oracle capable of quantifying how changes in model design at the microscopic level -- choice of computational primitives, parametrization, topology -- propagate to the macroscopic scale i.e., scaling laws. Indeed, a key challenge in architecture design is predicting whether new designs will match or improve quality of existing baselines at scale. 

Our working hypothesis is that the performance of an architecture primarily stems from its efficiency in performing an array of smaller token manipulation tasks well.
We show that by probing the performance of architectures in each of these individual tasks at a small scale, one can recover relative model rankings matching those obtained via scaling laws analysis in quantities of interest such as compute-optimal perplexity.

We call this process of capability identification and evaluation, with the goal of architecture prototyping, \textit{mechanistic architecture design} (in short "{\tt MAD}").
Beyond approximating scaling performance, {\tt MAD} provides a means to probe the compositionality of model skills.

\subsection{Synthetic tasks to probe model skills}
{\tt MAD} utilizes synthetic tasks to probe model skills and inform model design, building on recent works~\cite{fu2022hungry,poli2023hyena,arora2023zoology} considering only a single or subset of these tasks. We provide a schematic for each task, with $x$ representing the input, $y$ the target sequence, and {\tt prompt} the evaluation sequence.

\subsubsection{In-context recall}
\begin{wrapfigure}[11]{r}{0.45\linewidth}
    \centering
    \vspace{-3mm}
    \includegraphics[width=0.9\linewidth]{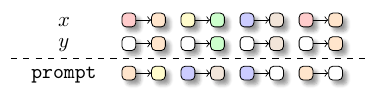}
    \caption{Schematic of in-context recall. White tokens are masked; $y$ represents target sequences during training. At test time, the model is evaluated on recall of all key-value pairs that were already presented in the sequence.}
    \label{fig:recall}
\end{wrapfigure}

To answer a prompt well, language models must be able to understand and learn from new information presented in the prompt (so-called in-context learning~\cite{elhage2021mathematical}).

A wealth of empirical work has demonstrated that the associative recall task, as studied in~\cite{fu2022hungry,poli2023hyena}, is well-suited to test a specific subset of in-context learning ability: direct lookup, requiring little to no processing of token embeddings to be solved \footnote{Solutions to in-context recall for some architectures can be expressed precisely and even hardcoded in an architecture without training \cite{massaroli2023laughing,arora2023zoology}. Thus, in-context recall tasks also represents a useful, albeit limited, test case to guide theoretical analysis.}. Here, we are using a multi-query variant of this task, as proposed by~\cite{arora2023zoology}: Given an input sequence of key-value pairs, models are tasked with retrieving all values from the input sequence associated with keys that were already shown in the input sequence.
Note that while the mapping from keys to values is consistent within an input sequence, it is randomly shuffled between sequences.

To solve this task, a model thereby does not need to learn any information external to the prompt it is provided with at test time.

\clearpage

\subsubsection{Fuzzy in-context recall}

\begin{wrapfigure}[9]{r}{0.45\linewidth}
    \centering
    \vspace{-3mm}
    \includegraphics[width=0.9\linewidth]{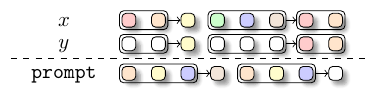}
    \label{fig:fuzzy-in-context}
    \caption{Fuzzy in-context recall. Boxes indicate adjacent tokens that form a key/value.}
\end{wrapfigure}

In language, semantic units are often spread out over multiple adjacent tokens (e.g., "blue sky" vs "gray sky"). To test how capable a model is of semantically grouping together adjacent tokens, we utilize a variant of in-context recall, in which keys and values are composed of a variable number of adjacent tokens.

For each sequence, variable length keys and values are randomly drawn from the vocabulary and then assigned into pairs.
Since the structure of key/value lengths in a sequence, as well as the mapping from keys to values, change between sequences, fuzzy recall can be regarded as a more challenging variant of in-context recall.

\subsubsection{Noisy in-context recall}

\begin{wrapfigure}[6]{r}{0.45\linewidth}
    \centering
    \vspace{-3mm}
    \includegraphics[width=0.9\linewidth]{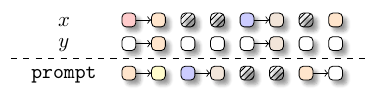}
    \label{fig:noisy-recall}
    \vspace{-4mm}
    \caption{Schematic of noisy in-context recall.}
\end{wrapfigure}

To answer a prompt well, language models must be able to ignore irrelevant information of the input.

We test this ability with another modification to standard in-context recall. Here, irrelevant information, represented by \textit{noise} tokens from a special subset of the vocabulary, is added in an arbitrary and variable pattern in between the key-value pairs.
Since the noise tokens are sampled from a fixed dictionary, this task  requires the model to implement a specific type of memory, in addition to the recall circuits required for in-context recall. In particular, the model needs to remember which tokens belong to the set of \textit{noise} tokens, as these do not carry relevant information for the task.

\vspace{2mm}
\subsubsection{Selective Copying}

\begin{wrapfigure}[9]{r}{0.45\linewidth}
    \centering
    \vspace{-3mm}
    \includegraphics[width=0.9\linewidth]{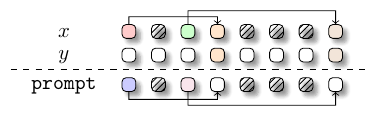}
    \label{fig:selective}
    \caption{Schematic of the selective copy task. Grayed-out tokens are noise.}
\end{wrapfigure}

In addition to ignoring irrelevant information of an input, language models must be able to selectively remember relevant information of an input.

In the selective copying task, models are tasked with copying tokens from one position of an input sequence to a later position of the sequence, while ignoring irrelevant noise tokens that are inserted into the sequence. Tokens are always copied in their order of occurrence. Models thereby need to not just remember the tokens that are to be copied but also their specific order of occurrence in the sequence. The copy positions are gleaned from the structure of each sample, while the contents change between samples and must be inferred in-context.

\subsubsection{Compression}
\begin{wrapfigure}[10]{r}{0.45\linewidth}
    \centering
    \vspace{-3mm}
    \includegraphics[width=0.82\linewidth]{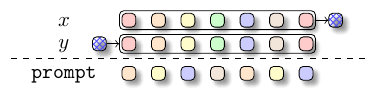}
    \label{fig:ecompression}
    \vspace{-1mm}
    \caption{Schematic of the compression task. A sequence is encoded into a single token, and then decoded to reconstruct the original sequence.}
\end{wrapfigure}

Recent findings in the mechanistic interpretability literature~\cite{nanda2023fact} indicate that language models are often required to perform "token concatenation", where early sequence-mixing layers (e.g., attention) assemble information that is spread across multiple tokens in an input onto another token so that the assembled information can then be decoded well by subsequent channel-mixing layers (e.g., MLPs). 

To test this capability we use a compression task, in which models are tasked with compressing a random sequence of input tokens into a single aggregation token, in a way that enables reconstruction via an MLP. In other words, the compression task tests the ability of a model to compress token embeddings into a single one with the least amount of information loss. 

\clearpage

\subsubsection{Memorization}
\begin{wrapfigure}[8]{r}{0.45\linewidth}
    \centering
    \vspace{-5mm}
    \includegraphics[width=0.75\linewidth]{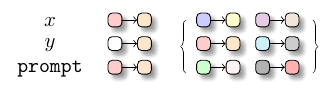}
    \label{fig:enter-label}
    \caption{Schematic of the memorization task. The model is tasked with learning a fixed map between tokens (i.e., a set of ``facts''). }
\end{wrapfigure}

In addition to manipulating and retrieving information from an input sequence, language modeling requires the memorization of factual knowledge.

To test this skill, we utilize a memorization task, in which models are tasked with learning a fixed key-value mapping (resembling facts in language) from the training data. Unlike recall, the mapping requires no in-context computation as the ground-truth mapping is constant across samples.

\subsection{{\tt MAD} Protocol}
{\tt MAD} follows a two-step procedure, starting from the design of a new candidate architecture, followed by its systematic evaluation according to the following key principles: 
\begin{itemize}[leftmargin=0.3cm]
    \item[$i.$] Each {\tt MAD} score is obtained by averaging architecture performances across a range of task difficulty levels. To manipulate difficulty, we independently vary a set of relevant experimental variables: length of the input sequence, size of the vocabulary, and size of the training set. Some tasks have additional variables such as the ratio of noise tokens in the noisy recall and selective copying tasks (Appendix~\ref{appendix:mad-tasks} and~\ref{appendix:mad-results}).
    \item[$ii.$] Fixed-state architectures are normalized to an iso-state and iso-parameter setting, including models featuring sparsely activated layers such as \textit{mixtures of experts} (MoEs)~\cite{shazeer2017outrageously}. Here, we normalize all fixed-state architectures to a common total state dimension of $4096$ to control for any differences in model performance driven primarily by mismatch in model state dimension (Appendix~\ref{appendix:mad-architectures}).
    \item[$iii.$] To ensure that model performance estimates are not dependent on a specific training setting, we sweep each architecture in each task setting over a grid of learning rate and weight decay values. We only include the best runs in our final analysis (Appendix~\ref{appendix:mad-training}).
    \item[$iv.$] Model performances are always evaluated in an independent evaluation dataset, specific to each task setting.
\end{itemize}

An implementation of the {\tt MAD} tasks are available at \url{https:/github.com/athms/mad-lab}.

\subsection{Candidate architecture designs}

We apply {\tt MAD} to a set of small two-blocks architectures built from a collection of common primitives such as attention, SwiGLU \cite{shazeer2020glu}, and variants of efficient implicit recurrent and convolutional layers described in Sec. \ref{taxonomy}.
We build different types of architectures with these primitives:  sequential, striped, and sparse parallel (mixtures).

In total, we evaluate $21$ distinct architectures, including combinations of the primitives described in Sec. \ref{sec:background}. Additional architecture details are provided in (Appendix~\ref{appendix:mad-details}).

\paragraph{Mixture of Sequence Experts}

We further introduce to our ${\tt MAD}$ analysis a layer inspired by sparsely gated channel mixers, the \textit{Hyena experts} layer. In a Hyena experts layer with $E$ experts and $K$ active experts, a router selects from a set of smaller Hyena mixers, using a router $G(u): u\mapsto s$ from input sequence $u\in R^{T \times D} $ to scores $s \in R^{T \times K}$, defined as
\[
s_t = {\sf softmax}({\sf top}_K(u_t {\sf W}_g)), \quad {\sf W}_g \in \R^{D \times E},\quad t = 1, \dots, L 
\]
resulting in 
\[
{\sf HyenaExperts}(u)_t = \sum_{k'=1}^k s_{tk'} {\sf Hyena}(u)_{tk'}.
\]
An advantage of the Hyena experts layer is that only a subset of the total state dimension is used to compose the output at each time step. We note that sparsely gated recurrences have also been explored for recurrences in \cite{ren2024sparse}, and that other similar schemes for sparse gating at the state level are also possible using input-varying recurrent primitives.
\begin{figure*}[!t]
    \centering
    \includegraphics[width=\textwidth]{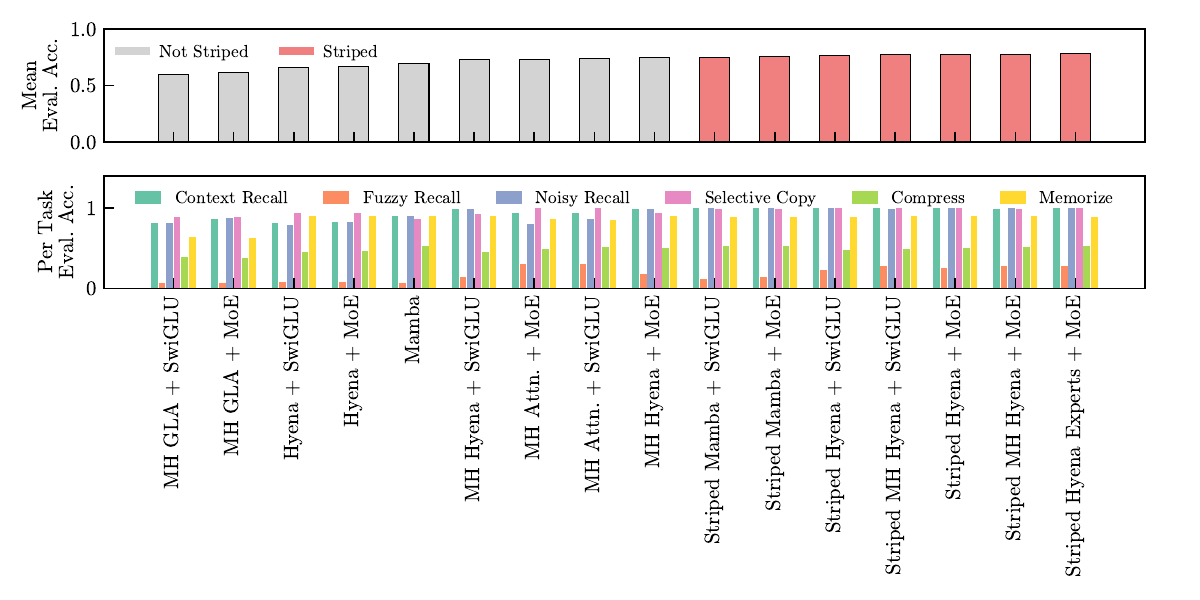}
    \vspace{-0.35in}
    \caption{{\tt MAD} analysis: An extensive evaluation of a suite of model architectures, built from common sequence- and channel-mixing layer types, across six synthetic tasks, each designed to probe a specific skill relevant for sequence modeling at scale.}
    \label{fig:mad-scores}
\end{figure*}

\subsection{Results}

We test a suite of architectures in the {\tt MAD} protocol.
In addition to ranking overall model performances across the synthetic tasks (Fig.~\ref{fig:mad-scores}), we take a high-level view on general patterns in model performances related to their design, including the presence of specific computational primitives in an architecture and the architecture's topology.
We indicate a model's performance by its accuracy in correctly predicting tokens in the synthetic tasks.
Note that model performances in {\tt MAD} can likewise be measured through their evaluation loss (see Appendix~\ref{appendix:fig:mad-scores-per-task}).
Both performance metrics yield similar model rankings.

\paragraph{Hybridization to combine specialized layers} 

Inspecting the performance on individual tasks via a stratified analysis (Appendix \ref{appendix:mad-results}) reveals specialization of architectures built with a single type of primitive, such as Mamba excelling at compression and Hyena at fuzzy recall.

\begin{tcolorbox}[enhanced, frame hidden, sharp corners, colback=blue!5, boxsep=0pt, before skip=0pt, after skip=0pt]
   \textbf{Finding $1$}: Striped architectures outperform all non-striped architectures on composite metrics, with an average gain in accuracy of $8.1\%$ across the {\tt MAD} synthetic tasks (Fig.~\ref{fig:mad-scores}). 
\end{tcolorbox}

We further find {\tt MAD} performance to increases with models' total fixed state dimension, underscoring the importance of normalizing state dimensions when comparing model capabilities, further motivating a state-optimal scaling law analysis (Fig. \ref{fig:state-scaling}). 

\paragraph{Head expansion trick} 

It is beneficial to arrange the fixed state dimension into larger heads with fewer states instead of smaller heads with additional states (in the limit case, in a mSISO configuration).

\begin{tcolorbox}[enhanced, frame hidden, sharp corners, colback=blue!5, boxsep=0pt, before skip=0pt, after skip=0pt]
   \textbf{Finding $2$}: Architectures that expand their total state dimension through heads (see Sec. \ref{taxonomy})
   outperform architectures without heads, with an average gain of $2.3\%$ in accuracy across the {\tt MAD} synthetic tasks (Fig.~\ref{fig:mad-scores}). 
\end{tcolorbox}

We note that the head expansion trick also linearly increases the computation in the layer, and for this reason it introduces a trade-off between compute-optimality and state-optimality.

In Sec. \ref{sec:results:scaling}, we will explore the trade-offs of this state configuration by comparing compute-optimal and state-optimal scaling of models with and without heads.
 
\paragraph{Sparse layers} 
We find sparsely gated layers to outperform dense layers in {\tt MAD} synthetics, in line with the literature on mixture of experts and their benefits.
\begin{tcolorbox}[enhanced, frame hidden, sharp corners, colback=blue!5, boxsep=0pt, before skip=0pt, after skip=0pt]
   \textbf{Finding $3$}: {\tt MAD} performance improves with the addition of sparsely activated mixture of expert channel-mixing layers, when compared to architectures using SwiGLU channel mixers, with an average gain in accuracy of $1.7\%$ across tasks  (Fig.~\ref{fig:mad-scores}). 
\end{tcolorbox}
In our later analyses, we will connect the performance of architectures on {\tt MAD} to their performance at scale on The Pile~\cite{gao2020pile} (Fig.~\ref{fig:mad-to-scale}).
Additional {\tt MAD} analysis results are provided in Appendix~\ref{appendix:mad-results}.

\section{Scaling Analysis}
\label{sec:results:scaling}

\begin{figure*}[!b]
    \centering
    \includegraphics[width=1.1\textwidth]{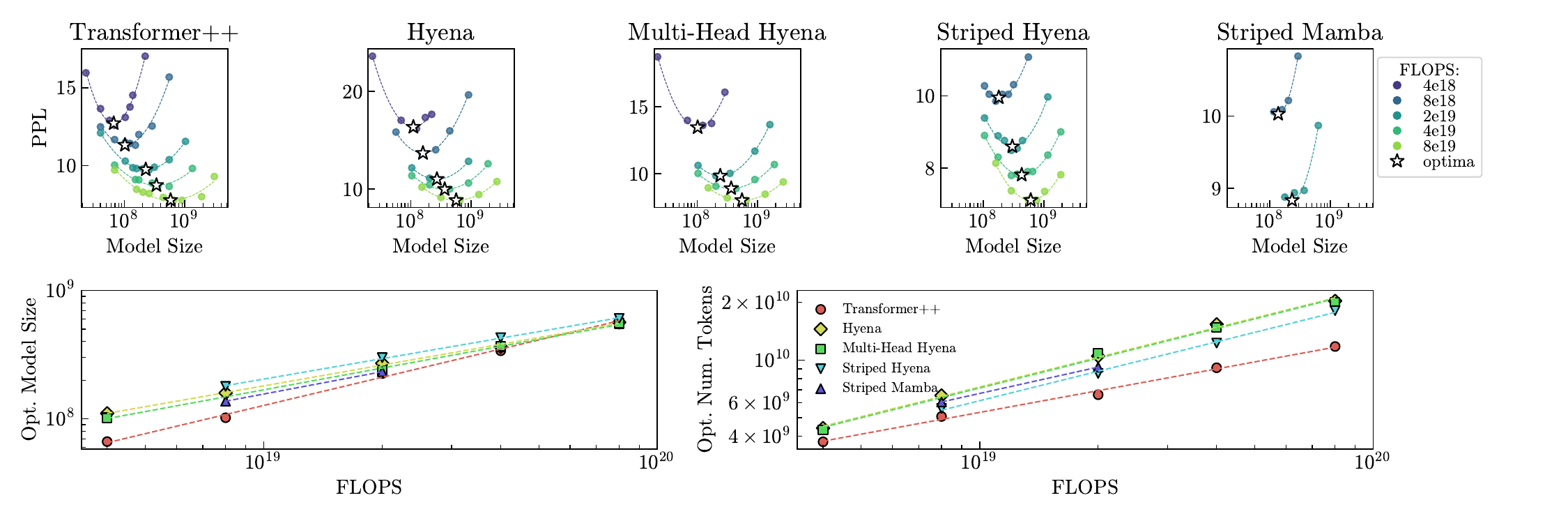}
    \vspace{-0.3in}
    \caption{Compute optimal scaling. \textbf{[Top:]} For each architecture, we train models of different sizes for a constant number of FLOPs (so-called IsoFLOP groups). For each of these IsoFLOP groups, we determine an optimum model size based on a polynomial fit to the observed training perplexities. \textbf{[Bottom:]} Using these estimates, we predict optimal model sizes and number of training tokens for each architecture.}
    \label{fig:efficient-frontier}
\end{figure*}

We seek to verify the connection between mechanistic design tasks and performance at scale. For this reason, we execute an extensive scaling law analysis on language pretraining, expanding on the framework of~\cite{kaplan2020scaling,hoffmann2022training}.
We train more than $500$ models of different architectures on The Pile \cite{gao2020pile}.

Let $\cM_{w, \xi}$ be a model with parameters $w$ and architecture $\xi$. Denote with $N = |w|$ the number of parameters, with $D$ the total number of training tokens, and the training cost (in \textit{floating point operations}, FLOPS) with $c_{\xi}(N, D)$.
Let ${\cal{A}}_{\xi}(C)$ be the set of tuples $(N, D)$ such that the training cost is exactly $C$, ${\cal{A}}_{\xi}(C) \coloneqq \{(N, D)~|~c_{\xi}(N, D)=C\}$. Given a tuple $(N,D)\in{\cal{A}}_{\xi}(C)$ one can evaluate $\cL_\xi(N,D)$, the loss achievable for that combination of parameters/tokens.
A point $(C, \ell(C))$ in the locus of the \textit{compute-optimal} frontier in the loss-compute plane is defined as
\[
    (C, \ell(C))~:~ \ell(C) = \min_{(N,D)\in{\cal{A}}_{\xi}(C)}\cL_{\xi}(N, D)
\]


with $\ell(C)$ indicating the best loss achievable by training $\cM_{\theta, \xi}$ at compute budget $C$, optimizing the allocation of compute to model size $N$ and training tokens $D$, for architecture $\xi$. Relatedly, one may seek the functional form of the compute-optimal frontier in the parameter-compute or token-compute planes, composed of tuples $(C, N^*)$ and $(C, D^*)$, where $D^*, N^*$ represent the optimal i.e., achieving lowest loss, allocation subject to the $(N^*, D^*) \in \cal{A}_{\xi}(C)$ constraint.

A primary objective of scaling law analyses is to determine such optimal allocation of the computational budget. To estimate efficient frontiers, we use an IsoFLOP approach, which explores different allocation ratios of model parameters and number of tokens at each compute budget. The loss optimum is then estimated via a quadratic fit (see Fig \ref{fig:optimal-stripes} as an example).

\subsection{Compute-optimal frontier for new architectures}

Our first set of findings is related to the efficient frontier of the baseline Transformer{\tt ++} \cite{touvron2023llama} in relation to other architectures. \cite{hoffmann2022training} finds that when $\xi$ is a standard Transformer architecture (combining attention and MLP), the optimal ratios between the number or model parameters, training tokens, and compute budget, are explained by a linear relationship in log-log space, i.e., $\log N^* \propto a \log C$ and $\log D^* \propto b \log C$.

\begin{tcolorbox}[enhanced, frame hidden, sharp corners, colback=blue!5, boxsep=2pt, before skip=1pt, after skip=1pt]
   \textbf{Finding $5$}: Let $a_{\tt H}, a_{\tt T}, b_{\tt H}, b_{\tt T}$ be the parameter size and data allocation coefficients for striped and Transformer models, respectively. We estimate $a_{\tt T} > a_{H}$ and $b_{\tt T} < b_{H}$ (Fig. \ref{fig:efficient-frontier}).
\end{tcolorbox}
Optimal allocation of tokens and parameters is relatively stable under striping, with marginal differences. One notable difference is that optimal compute allocation in emerging efficient architectures is skewed towards additional data i.e., training smaller models for longer.

\begin{figure*}[h!]
    \centering
    \includegraphics[width=\textwidth]{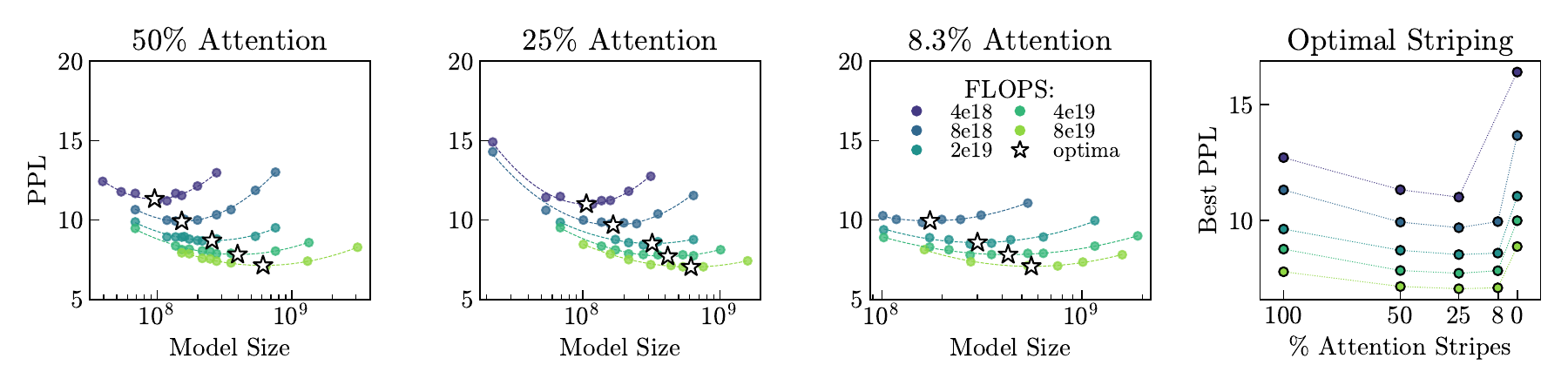}
    \vspace{-6mm}
    \caption{Optimal striping ratio. We find that StripedHyena architectures outperform non-striped Hyena (0\% Attention) and Transformer{\tt ++} (100\% Attention) architectures across all evaluated FLOPS groups. In particular, we find a ratio of $25\%$ to be optimal.}
    \label{fig:optimal-stripes}
\end{figure*}

\paragraph{Beyond the efficient frontier}
Next, we look at optimality gaps when training outside the efficient frontier. By optimality gap, we refer to the increase in loss by training outside the compute-optimal frontier i.e., $\mathcal{L}(C(\tilde{N},\tilde{D},\xi))$ where $\tilde{N} = N^* + \delta N^*$ and the number of tokens $\tilde{D}$ is adjusted to preserve the total compute cost. 
\begin{tcolorbox}[enhanced, frame hidden, sharp corners, colback=blue!5, boxsep=2pt, before skip=1pt, after skip=1pt]
   \textbf{Finding $6$}: The off compute-optimal perplexity gap is proportional to the hybridization ratio (Fig.\ref{fig:optimal-stripes}), for all IsoFLOP groups.
\end{tcolorbox}
Intuitively, models with "flatter" IsoFLOP perplexity curves are preferred for overtraining smaller models, a setting particularly common in practice, as it results in smaller models with faster inference. Interestingly, the suboptimality gap in hybrids is smaller than Transformers, meaning they are better suited to training outside the optimal frontier.
\paragraph{Striping schedule and topology}
We study compute-optimal ratio and allocation of attention operators in striped architectures, as well as their overall topology (Fig. \ref{table:striping-topology}).

\begin{tcolorbox}[enhanced, frame hidden, sharp corners, colback=blue!5, boxsep=2pt, before skip=1pt, after skip=1pt]
   \textbf{Finding $7$}: The compute-optimal hybridization ratio for striped models is $25\%$ across all IsoFLOP groups\footnotemark (Fig.\ref{fig:optimal-stripes} and Table~\ref{table:striping-topology}).
\end{tcolorbox}
\footnotetext{Accounting for state-optimality shifts the optimal ratio to $10\%$.}

\begin{figure}[t]
    \centering
    \includegraphics[width=0.99\linewidth]{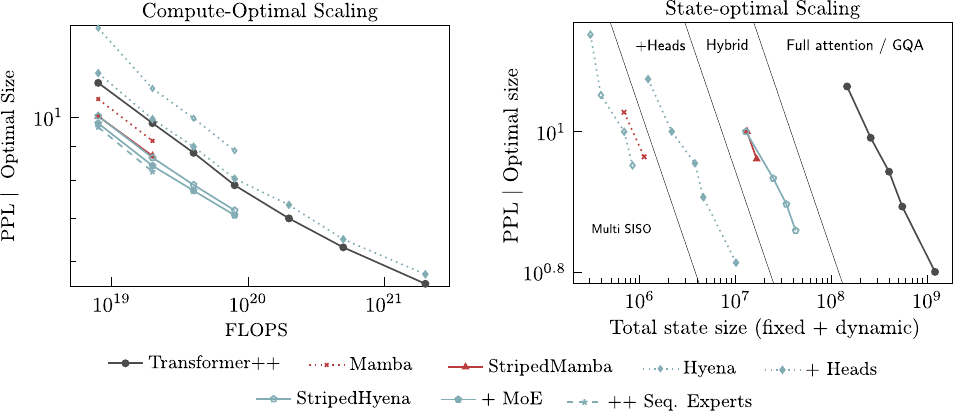}
    \caption{Compute-optimal and state-optimal scaling on The Pile. We report total state dimension, fixed (recurrences) and dynamic (attention). All models are trained at sequence length 8k. We identify distinct regions in the state-optimal frontier, indicating that one may pay an additional FLOP cost to obtain the same perplexity with a state of smaller dimension, by using other classes of architectures.}
    \label{fig:state-scaling}
\end{figure}

\paragraph{Batch sizes and hyperparameters}

Batch size and learning rate are two high-impact hyperparameters for scaling laws, as they visibly shift the compute-efficient frontier. We find that scaling the batch size with FLOP budgets, thus keeping it fixed within each IsoFLOP group, to be a simple and robust approach. Fig.~\ref{fig:batch-size-effect} provides an example of potential issues arising from incorrect batch scaling. These results are in line with recent findings \cite{bi2024deepseek}.  

\subsection{State-optimal scaling}
Beyond driving {\tt MAD} synthetics performance, the total state size in a model is also an important factor in determining inference latency and memory cost. We explore \textit{state-optimal} scaling, aiming to provide a coarse estimate of state utilization by measuring scaling in perplexity over state dimension (Fig. \ref{fig:state-scaling}, right). 

\begin{tcolorbox}[enhanced, frame hidden, sharp corners, colback=blue!5, boxsep=2pt, before skip=1pt, after skip=1pt]
   \textbf{Finding $8$}: There exists a relation of the type $P^* \propto M^c$ between compute-optimal perplexity $P^*$ and total state size $M$, with $c \approx -0.28$ in our scaling experimental setup, consistent across all model architectures. The model class determines the offset of the state-optimal curve. 
\end{tcolorbox}

Concretely, state-optimal scaling indicates that one may reach any target perplexity (up to saturation of compute-optimal scaling laws i.e., approaching entropy of text) with fixed-state architectures, by paying a FLOP cost multiplier that depends on the model class -- training longer to maximize state utilization. Input-varying recurrences, multihead and striped hybrid architectures achieve a favourable trade-off between metrics, with comparable or improved compute-optimal perplexity to Transformers{\tt ++} and a reduced total state dimension.


%
\subsection{Compute-optimal scaling at byte resolution}

\begin{wrapfigure}[13]{r}{0.35\linewidth}
    \centering
    \vspace{-1mm}
    \includegraphics[width=\linewidth]{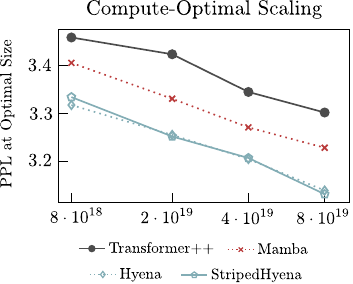}
    \vspace{-6mm}
    \caption{Compute-optimal scaling at byte resolution.}
    \label{fig:dna-compute-optimal}
\end{wrapfigure}

Scaling laws analysis primarily focus on sub-word level tokenization. With a new range of architectural options, we also explore compute-optimal scaling of a subset of architectures (Transformer{\tt ++}, Mamba, Hyena and StripedHyena) at byte resolution.
We scale the models across FLOP budgets from 8e18 to 8e19 with model sizes from 6M to 1B parameters. The compute-optimal frontier is obtained using a similar protocol as outlined in Sec. \ref{appendix:scaling-laws}, with additional details and results shown in Sec. \ref{appendix:byte}.

We find attention-based models to yield significantly higher perplexity at all IsoFLOP groups, with alternative architectures outperforming Transformer{\tt ++}, including non-striped variants (Figure \ref{fig:dna-compute-optimal}). These results show that model ranking varies significantly across domains and tokenization strategies.

\section{Connecting {\tt MAD} to scaling metrics}

The goal of {\tt MAD} is to provide a framework that can accelerate the architecture design process by using small synthetic tasks, which can be evaluated quickly and with little compute, to estimate whether improvements to an existing architecture, or a new candidate architecture, will perform well at scale. To gauge this hypothesis, we study the correlation between {\tt MAD} scores and scaling properties of interest.

\paragraph{Correlation to compute-optimal perplexity}

\begin{figure}[!t]
    \centering
    \vspace{-0.2in}
    \includegraphics[width=\columnwidth]{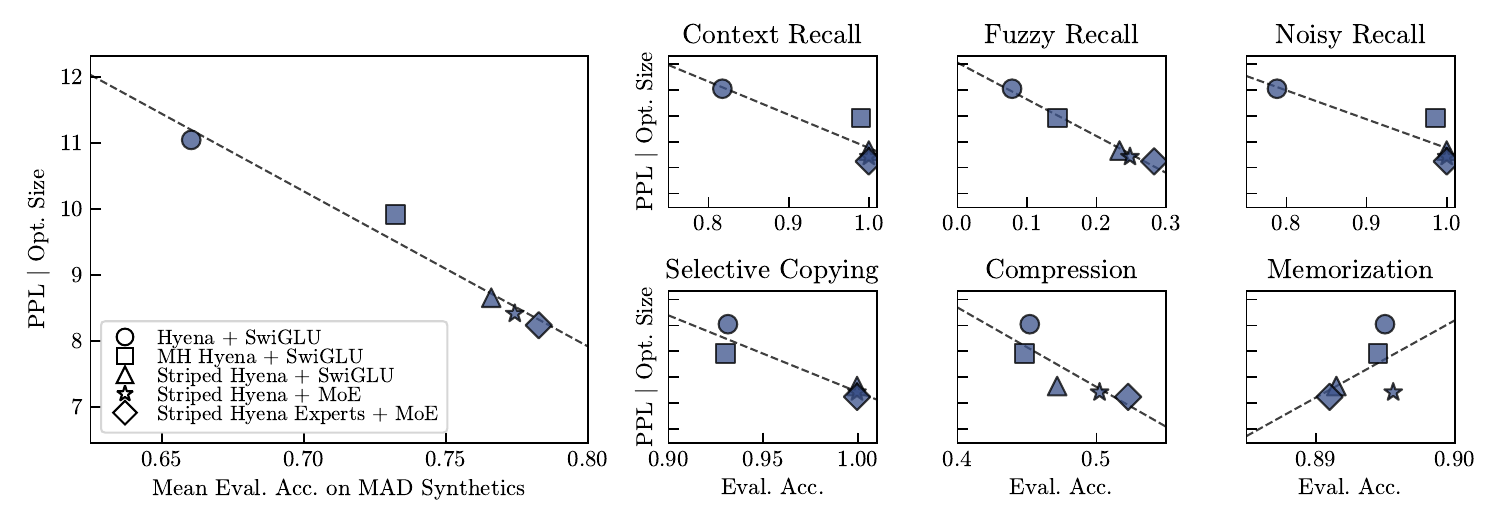}
    \vspace{-0.15in}
    \caption{Improved performance on {\tt MAD} synthetics correlates with better compute-optimal perplexity on The Pile. We highlight progressively improved versions of Hyena that were designed with the {\tt MAD} pipeline, which translated to improved perplexity on the Pile (shown for $2e19$ FLOPs; see Appendix~\ref{appendix:fig:mad-to-scale-hyena-full} for an analysis across IsoFLOP groups).}
    \vspace{-4mm}
    \label{fig:mad-to-scale}
\end{figure}

We start with a case study using the Hyena~\cite{poli2023hyena} architecture.
{\tt MAD} has indicated that the performance of Hyena can be cumulatively improved by i) adding heads to the Hyena sequence mixer, ii) interleaving Hyena and attention layers, iii) using a sparse MoE channel mixer instead of SwiGLU, and iv) integrating a sparse routing mechanism into the Hyena sequence mixer (Fig.~\ref{fig:mad-scores}).
Using the results of our scaling analysis (Sec.~\ref{sec:results:scaling}), we can investigate the correlation between the {\tt MAD} scores of these architectures, as indicated by their average accuracy across the synthetic tasks, and their compute-optimal performance on The Pile (Fig.~\ref{fig:mad-to-scale} left). We also consider perplexity on {\tt MAD} tasks as an additional metric (Appendix \ref{appendix:mad-results}).
\begin{tcolorbox}[enhanced, frame hidden, sharp corners, colback=blue!5, boxsep=2pt, before skip=1pt, after skip=1pt]
   \textbf{Finding $9$}: Aggregate {\tt MAD} scores are linearly correlated with compute-optimal perplexity at scale for all compute budgets (Fig. \ref{fig:mad-to-scale} left, Appendix \ref{appendix:fig:mad-to-scale-hyena-full}).
\end{tcolorbox}

This result suggests that smaller, shallower models unit tested on {\tt MAD} synthetics can be used to predict compute-optimal scaling, as well as to iterate on improvements to a base architecture. To better understand the contribution of each {\tt MAD} task to the predictive power of the scores, we also report correlation for single-task performances and compute-optimal perplexity at scale (Fig. \ref{fig:mad-to-scale} right). We note that there is very little variation in the architectures' performances on the memorization task, which could explain why we did not find an association between their performances on this task and their performances at scale.

Next, we replicate this analysis for the Mamba architecture~\cite{gu2023mamba}, comparing the base architecture to a striped hybrid variant (Appendix~\ref{appendix:fig:mad-to-scale-mamba}).
Again, improved performances on {\tt MAD} correlate with improved compute-optimal perplexity on The Pile, underlining the generalizability of {\tt MAD}. Correlation across different architecture classes, although present (see Fig. \ref{fig:catchy}), is subject to more noise. Improvements to the pipeline and selection of evaluation settings {\tt MAD} may be required to minimize the impact of spurious hyperparameters.

\paragraph{Extensions and limitations}

The {\tt MAD} evaluation framework relies on extrapolating performance from smaller (e.g., 2-block) models to deeper models trained at scale.
As such, the framework has not yet been applied to sophisticated topologies requiring small-scale testing with a larger number of blocks e.g., hybrid models with more than two sequence mixer primitives, or alternative interconnection topologies that span multiple layers.

In principle, {\tt MAD} can be used to design architectures to optimize other quantities of interest, beyond perplexity or downstream benchmarks e.g., throughput.
In this work, we focus on investigating correlation with compute-optimal scaling metrics, and leave other analyses to future work.
\section{Conclusion}

This work explores architecture optimization, from synthetic tasks designed to probe specific model capabilities to scaling laws. We introduce \textit{mechanistic architecture design} ({\tt MAD}), a methodology for fast prototyping and verification of new deep learning architectures based on key token manipulation tasks such as recall and compression. With {\tt MAD}, we identify hybridization and new configurations to improve compute-optimal scaling of new architectures. We carry out an extensive scaling law analysis of new architectures, training over $500$ models between parameter sizes of $70$M to $7$B, verifying the improvements found via {\tt MAD}, and derive a collection of novel insights on the optimal scaling of new architectures. We introduce state-optimal scaling as a measure of efficiency for blocks with a fixed-size state, with implications for inference memory and latency. Finally, we show how {\tt MAD} results are correlated with perplexity in a compute-optimal regime, paving the way for faster and cheaper architecture prototyping. Overall, this work provides evidence of correlation between scaling and a selection of synthetic token manipulation tasks, as well as of the existence of a variety of hybrid architectures improving over Transformers at scale and on individual tasks.

\section{Ethical Impact}
This paper introduces \textit{mechanistic architecture design} ({\tt MAD}), a methodology for improving the scaling performance of deep learning models, and presents several improved architectures. As a consequence of this line of work, we expect training and inference of large models to become more efficient, less expensive, and thus more readily available. Societal consequences related to the existence of large foundation models based on Transformers also apply when discussing new improved architectures.

\section{Acknowledgments}
We are grateful to the Hessian.AISC Service Center, funded by the Federal Ministry of Education and Research (BMBF), for the collaboration and joint use of their supercomputer forty-two.
\printbibliography
\newpage
\clearpage

\appendix

\rule[0pt]{\columnwidth}{1pt}
\begin{center}
    \huge{\sc Mechanistic Design and Scaling of Hybrid Architectures} \\
    \vspace{0.15cm}
    \emph{Supplementary Material}
\end{center}
\rule[0pt]{\columnwidth}{1.5pt}

\doparttoc
\tableofcontents

\section{Additional Related Work}

\paragraph{Synthetics for analysis and design}
The {\tt MAD} framework builds on work on synthetic tasks for mechanistic interpretability of RNNs and Transformers, including associative recall, reasoning tasks, compression. \cite{olsson2022context} and a number of follow up in mechanistic interpretability use an induction task to probe into the internals of Transformer model. There is a large body of work \cite{weiss2018practical,hewitt2020rnns} studying the expressivity of recurrent models, either theoretically or empirically, using formal languages and other token manipulation tasks. 

Smaller scale synthetics have been used during the iterative design procedure of new layers and primitives, particularly in the context of emerging deep signal processing architecture. \cite{dupont2019augmented,massaroli2020dissecting,gu2021efficiently,fu2022hungry,zhang2023effectively,poli2023hyena,arora2023zoology}. Notably, \cite{fu2022hungry} uses associative recall to identify a key capability gap in previous gated state-space models, and proposes a modification to the layer. \cite{poli2023hyena} extend associative recall procedure to longer sequences, introducing new synthetic tasks such as counting. However, the pretraining results only involve smaller models, and are not obtained via compute-optimal scaling. 

There exists a long line of work on neural architecture search methods (see \cite{white2023neural} for a review). {\tt MAD} provides a different approach based on synthetic tasks. {\tt MAD} metrics are in principle compatible with various search methods.

\paragraph{Synthetics for evaluation}

Synthetics have also been leveraged to evaluate models and model classes
\cite{arora2023zoology,bhattamishra2023understanding,akyurek2024context}. \cite{poli2023hyena} shows correlation between synthetics and pretraining results on The Pile. \cite{arora2023zoology} maps associative recall accuracy gaps to a perplexity gap between pretrained models. A variety of other analyses on synthetics for emerging architectures finds certain classes of efficient architectures to be on par or outperform Transformers on most tasks, with gaps on tasks involving heavy recall or copying of tokens. With {\tt MAD}, we aim to leverage tasks as unit tests with a quantitative connection to scaling properties, instead of using smaller-scale experiments to only build intuition on potential model differences. 

\paragraph{Scaling laws}

We extend the compute-optimal scaling law analysis protocol of \cite{kaplan2020scaling,hoffmann2022training} performed on Transformers to deep signal processing architectures, including hybrids and sparsely gated architectures. We base the scaling analysis in this work on the compute-optimal protocol, in order to evaluate relative performance and to identify optimal hybridization ratios. Moreover, we consider extensions such as state-optimal scaling and performance in overtrained regimes (outside the compute-optimal frontier), both of which have implications for efficient inference.

\begin{figure}[!h]
    \centering
    \begin{subfigure}[b]{0.45\textwidth}
        \centering
        \includegraphics[width=0.8\textwidth]{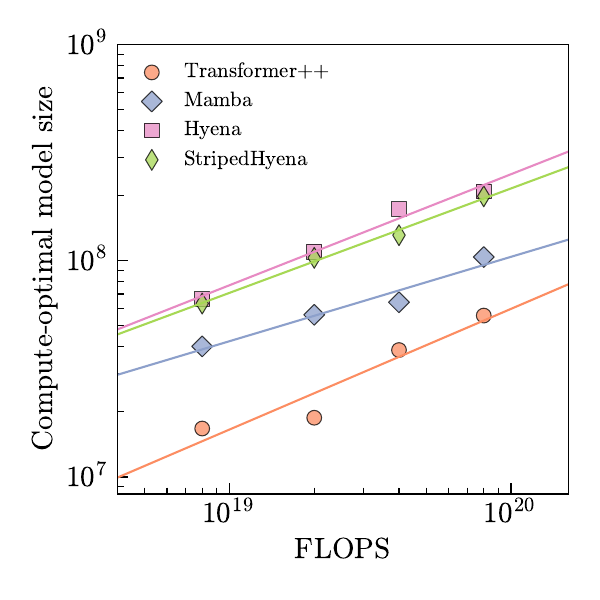}
        \caption{Optimal model size vs FLOPS.}
        \label{fig:dna-model-size}
    \end{subfigure}
    \hfill
    \begin{subfigure}[b]{0.45\textwidth}
        \centering
        \includegraphics[width=0.8\textwidth]{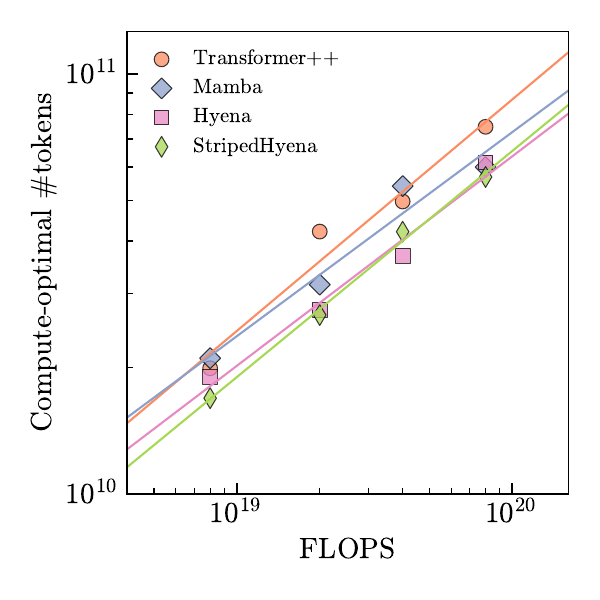}
        \caption{Optimal number of tokens vs FLOPS.}
        \label{fig:dna-tokens}
    \end{subfigure}
    \hfill
    \caption{Comparison of optimal model size and number of tokens for each FLOP budget.}
    \label{fig:dna-comparison}
\end{figure}

Other work on evaluation of new architectures experiments in parameter-matched and data-matched regimes, which can result in a mismatch with scaling results due to different FLOP costs per iteration. Other notable examples of compute-matched evaluations for new models are provided in \cite{poli2023hyena,gu2023mamba}. Previous evaluations are not carried out at compute-optimal model sizes which can vary significantly across architectures see e.g., Figures \ref{fig:efficient-frontier} and \ref{fig:dna-model-size}).

\section{Mechanistic Architecture Design}
\label{appendix:mad-details}

\subsection{Tasks}
\label{appendix:mad-tasks}

\subsubsection{In-Context Recall}
The in-context recall task is comprised of sequences of key-value pairs (with separate vocabularies for keys and values).
Models are tasked with predicting all values for those keys that were already presented in the sequence:

\begin{tcolorbox}[enhanced, frame hidden, sharp corners, colback=gray!15, boxsep=2pt, before skip=1pt, after skip=1pt]
    \begin{center}
        input example: {\tt a} {\tt b} {\tt d} {\tt e} {\tt f} {\tt g} | {\tt a} \underline{{\tt b}} {\tt f} \underline{{\tt g}}
    \end{center}
\end{tcolorbox}

In this example, keys are drawn from the vocabulary \{a, d, f\} 
and values from the \{b, e, g\} vocabulary.
Importantly, the mapping from keys to values is randomly shuffled between sequences. 
Models are tasked with autoregressively predicting all underlined value in this example.

In the baseline setting of this task, we use a vocabulary of $16$ tokens and $12,800$ training sequences with a length of $128$ tokens. The vocabulary is equally divided into keys and values.

\subsubsection{Fuzzy In-Context Recall}
The fuzzy in-context recall tasks adapts the in-context recall task by representing keys and values by a variable number of adjacent tokens:

\begin{tcolorbox}[enhanced, frame hidden, sharp corners, colback=gray!15, boxsep=2pt, before skip=1pt, after skip=1pt]
    \begin{center}
        input example: {\tt (a d) (b) (d a f) (e g) | (d a f) (\underline{e} \underline{g})}
    \end{center}
\end{tcolorbox}

In this example, keys are drawn from the vocabulary \{a, d, f\} and values are drawn from the vocabulary \{b, e, g\}.
We use brackets for illustrative purposes to indicate adjacent tokens that together represent a key or value but they are not part of the actual input to the model. 
In sequential order, the presented keys are 'a d' and 'd a f', with associated values 'b' and 'e g'.
For each sequence, keys and values are randomly drawn form the key and value dictionaries, with randomly drawn lengths (ranging from $1$ to $3$ tokens in our analyses).
We always evaluate with keys of length $3$ (the longest length used in our analyses), to disambiguate whenever a key token appears in two keys of different values.
We pad sequences with a separate pad token if necessary to ensure that all sequences of a dataset are of the exact same length.
As for the in-context recall task, models are tasked with autoregressively predicting all underlined values in this example.

In the baseline setting of this task, we use a vocabulary of $16$ tokens and $12,800$ training sequences with a length of $128$ tokens. The vocabulary is equally divided into key and value tokens.

\subsubsection{Noisy In-Context Recall}
The noisy in-context recall task represents another variation of in-context recall, in which noise tokens, from a separate vocabulary, are randomly inserted into the input sequences:

\begin{tcolorbox}[enhanced, frame hidden, sharp corners, colback=gray!15, boxsep=2pt, before skip=1pt, after skip=1pt]
    \begin{center}
        input example: {\tt a b h d e f g | i a \underline{b} f \underline{g}}
    \end{center}
\end{tcolorbox}

In this example, keys and values are respectively drawn from the vocabularies \{a, d, f\} and \{b, e, g\}, while noise is drawn form the vocabulary \{h, i\}.
As for in-context recall, models are tasked with autoregressively predicting the underlined values in this example.

In the baseline setting of this task, we use a vocabulary of $16$ tokens, which are equally divided into keys and values, $12,800$ training sequences with a length of $128$ tokens, and a share of $20\%$ noise tokens in the input from a separate noise vocabulary of size $16$.

\subsubsection{Selective Copying}
The selective copying task comprises sequences of randomly sampled tokens, with randomly inserted "blank" and "insert" tokens:

\begin{tcolorbox}[enhanced, frame hidden, sharp corners, colback=gray!15, boxsep=2pt, before skip=1pt, after skip=1pt]
    \begin{center}
        input example: {\tt a c [b] t [b] [i] [i] [b] [i] | a c [b] t [b] \underline{a} \underline{c} [b] \underline{t}}
    \end{center}
\end{tcolorbox}

In this example, tokens are drawn from the vocabulary \{a,c,t\}, while [b] and [i] indicate the blank and insert token.
Given this example, the task of the model is to copy all non-special tokens to the positions of the insert tokens, in the order they were presented in the sequence.
The purpose of the randomly inserted blank tokens is to force models to learn to selectively memorize or ignore information from the input.

In the baseline setting of this task, models are tasked with copying $16$ randomly drawn tokens from a vocabulary of $16$ tokens, and are provided with $12,800$ training sequences with a length of $256$ tokens.

\subsubsection{Compression}
The compression task consists of random token sequences, each ending with a dedicated "compression token":

\begin{tcolorbox}[enhanced, frame hidden, sharp corners, colback=gray!15, boxsep=2pt, before skip=1pt, after skip=1pt]
    \begin{center}
        input example: {\tt a e c b h g i [c] | [c] + [$pos_0$] -> a}
    \end{center}
\end{tcolorbox}

In this example, tokens are randomly drawn from the vocabulary \{a, b, c, e, g, h, i\}, while [c] indicates the compression token.
Given this input, models are tasked with compressing all relevant sequence information into the compression token [c], such that a subsequent two-layer MLP can fully recover each token of the input sequence, given the model's output for the compression token.
To indicate the position $i$ that is to be recovered from the input, we add a non-learnable sin-cos position embedding (indicated by $[pos_i]$) to the models output for the compression token before feeding it to the MLP decoder.

In the baseline setting of this task, we use a vocabulary of $16$ tokens and $12,800$ training sequences with a length of $32$ tokens.

\subsubsection{Memorization}
The memorization task uses a fixed key-value dictionary, representing the facts to be learned:

\begin{tcolorbox}[enhanced, frame hidden, sharp corners, colback=gray!15, boxsep=2pt, before skip=1pt, after skip=1pt]
    \begin{center}
         key-value dictionary example: {\tt \{a:b, c:d, e:f\}}
    \end{center}
\end{tcolorbox}

Input sequences comprise key-value pairs that are randomly sampled form this dictionary.
Importantly, all values are masked out form the input sequences with a dedicated "insert token":

\begin{tcolorbox}[enhanced, frame hidden, sharp corners, colback=gray!15, boxsep=2pt, before skip=1pt, after skip=1pt]
    \begin{center}
         input example: {\tt a [i] c [i] e [i] a [i] | a \underline{b} c \underline{d} e \underline{f} a \underline{b}}
    \end{center}
\end{tcolorbox}

In this example, the values that are to be inserted at the positions of the insert tokens are: 'b', 'd', 'f', and 'b'.
Models are then tasked with correctly inserting the masked-out values at the positions of the insert tokens.
As the values are never part of the input sequences, models need to learn the mapping from keys to values over the course of their training.

In the baseline setting of this task, we use a vocabulary of $256$ tokens, equally divided into keys and values, and $256$ training sequences with a length of $32$ tokens (such that each fact is on average presented $32$ times in the training data).

\subsection{Manipulating Task Difficulty}
For each {\tt MAD} task, we evaluate model performances across several levels of difficulty.
We manipulate task difficulty by i) increasing the length of the input sequences, ii) reducing the training dataset size, and iii) increasing the vocabulary size.
In addition, we increase the share of noise in the inputs for the noisy in-context recall task as well as the number of tokens that are to be copied in the selective copying task.
Importantly, we only change one task variable at a time, while keeping all others at their baseline level.

For all variants of in-context recall, we evaluate input sequence lengths of $128$, $256$, $512$, and $1024$ tokens, training dataset sizes with $12,800$, $6,400$, $3,200$, $1,600$ and $800$ samples, and vocabulary sizes, which are equally divided into keys and values, of $16$, $32$, $64$, and $128$ tokens.

For noisy in-context recall, we additionally evaluate shares of $20\%$, $40\%$, $60\%$, and $80\%$ noise tokens in the inputs.

For the selective copying task, we evaluate sequence lengths of $256$, $512$, and $1024$ tokens, training dataset sizes with $12,800$, $6,400$, $3,200$, $1,600$ and $800$ samples, vocabulary sizes of $16$, $32$, $64$, and $128$ tokens, and $16$, $32$, $64$, and $96$ tokens of a the input that are to be copied.

For the compression task, we evaluate input sequence lengths of $32$, $64$, $128$ and $256$ tokens, vocabulary sizes of $16$, $32$, $64$, and $128$ tokens, and training dataset sizes of $12,800$, $6,400$, $3,200$, $1,600$ and $800$ samples.

For the memorization task, we evaluate vocabulary sizes of $256$, $512$, $1,024$, $2,048$, $4,096$, and $8,192$ tokens, while keeping the training dataset fixed at $256$ samples with an input length of $32$ (thereby effectively varying the rate at which each fact appears in the training data, with average rates of $32$, $16$, $8$, $4$, $2$, and $1$).

\subsection{Architectures}
\label{appendix:mad-architectures}

We build architectures from a set of common channel- and sequence-mixing layer primitives.
Each architecture is composed of $2$ blocks with a total of $4$ layers.
In general, blocks combine a sequence mixing layer with a subsequent channel mixing layer, with the exception of Mamba layers, which combine sequence and channel mixing into a single layer~\cite{gu2023mamba}.
All layers are set to a width of $128$ for our main analysis (if not stated otherwise), with all other architecture settings given below.

Common architecture primitives are composed of two identical blocks combining each sequence-mixing layer with each of the two channel-mixing layers.
Striped hybrid architectures combine each unique block of the common architecture primitives with a second block composed of multi-headed attention and one of the two channel mixers.

\subsubsection{Channel-mixing Layers}

\begin{itemize}
    \item \textbf{SwiGLU MLP~\cite{shazeer2020glu}:} inner width: $512$
    \item \textbf{Mixture of Experts MLP~\cite{lepikhin2020gshard}:} number of experts: $8$, expert width: $16$, number of active experts: $2$
\end{itemize}

\subsubsection{Sequence-mixing Layers}

We normalize the (fixed) state dimension of all sequence mixers, before running the MAD pipeline. Whenever possible, we prioritize keeping the shape of the layer fixed, over the state dimension (e.g., reducing state dimension before expansion factors, or  reducting state dimension before number of heads).

\begin{itemize}
    \item \textbf{Hyena~\cite{poli2023hyena}:} filter order: $2$, short filter order: $3$, filter featurization is implemented following \cite{massaroli2023laughing}.
    \item \textbf{Mamba~\cite{gu2023mamba}:} state dimension: 4, convolution dimension: $4$, width expansion: 2, no bias for linear and convolution layers.
    \item \textbf{Multi-head Gated Linear Attention~\cite{yang2023gated}:} number of heads: $8$, head dimension: $16$
    \item \textbf{Multi-Head Attention~\cite{vaswani2017attention}:} number of heads: $16$, head dimension: $8$, no bias for linear layers
    \item \textbf{Multi-Head Hyena~\cite{massaroli2023laughing}:} number of heads: $16$, state dimension of heads: $2$, filter order: $2$, short filter order: $3$.
    \item \textbf{Hyena Experts:} number of experts: $8$, expert width: $16$, number of active experts: $2$. All other parameters are shared with standard Hyena.
\end{itemize}

At these settings, all evaluated architectures that do not include attention layers are normalized to a total state dimension of $4,096$.

\subsection{Training}
\label{appendix:mad-training}

For each {\tt MAD} task, we train models according to the setting described in Table~\ref{table:mad-train-setting}, using a standard cross-entropy loss objective.
Note that we sweep all evaluated architectures over a $3 \times 2$ grid of learning rate and weight decay values (see Table~\ref{table:mad-train-setting}) and only include the best runs in our final analysis (as determined by their evaluation accuracy).

\begin{table}[!h]
\caption{{\tt MAD} training setting.}
\label{table:mad-train-setting}
\vskip 0.15in
\begin{center}
\begin{small}
\begin{sc}
\begin{tabular}{lcccr}
\toprule
Optimizer & AdamW\\
Optimizer momentum & $\beta_1,\beta_2=0.9, 0.98$\\
Dropout & None\\
Batch size & 128\\
Training epochs & 200\\
Learning rate schedule & cosine decay\\
Number of layers & 4\\
Number of evaluation samples & 1,280 \\
\midrule
Base learning rate & [$0.0001, 0.0005, 0.001$]\\
Weight decay & [$0.0, 0.1$]\\
\bottomrule
\end{tabular}
\end{sc}
\end{small}
\end{center}
\vskip -0.1in
\end{table}

\subsection{Results}
\label{appendix:mad-results}

\subsubsection{Task Performances}

\begin{figure*}[!h]
    \centering
    \vspace{-0.1in}
    \includegraphics[width=\columnwidth]{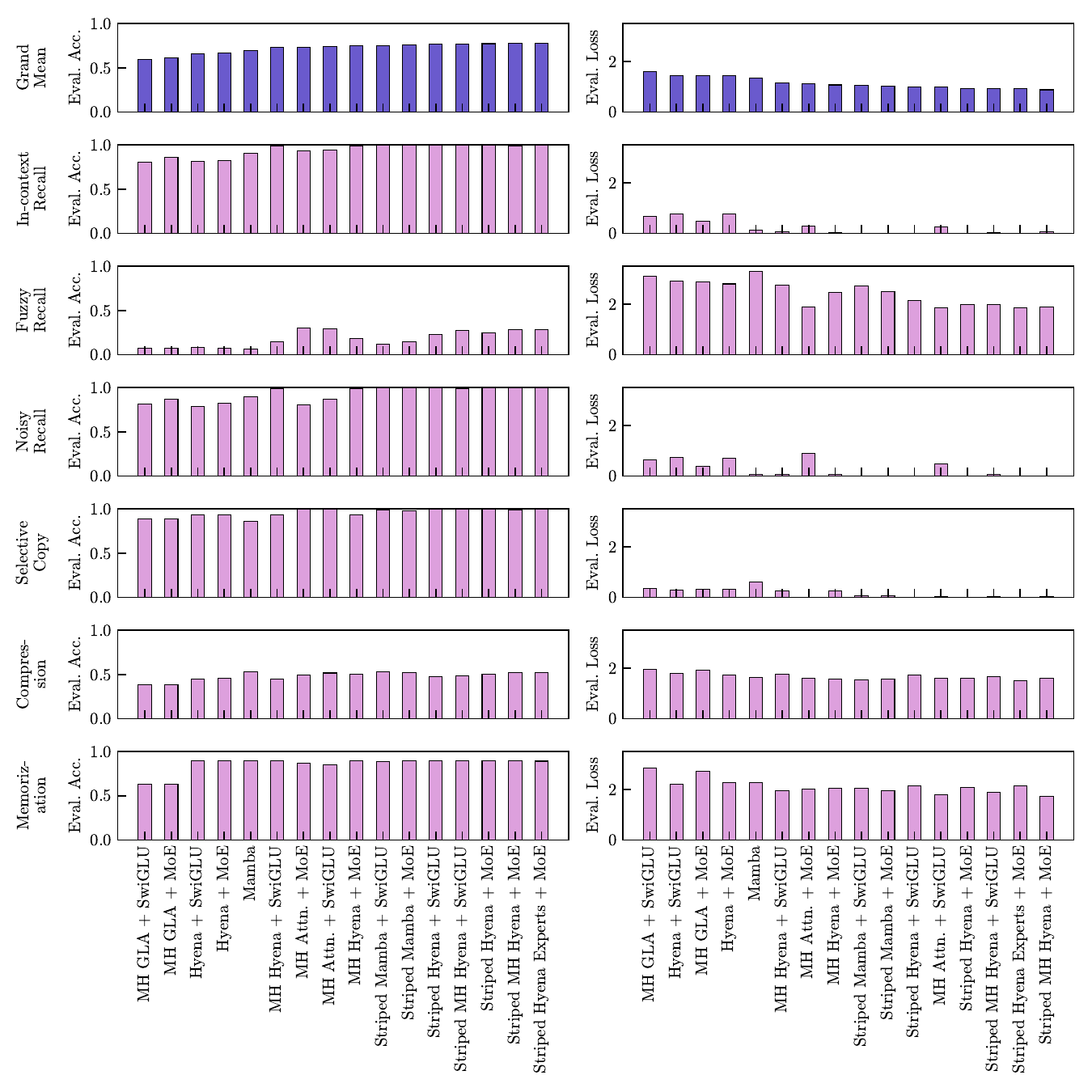}
    \vspace{-0.2in}
    \caption{Architecture performances within and across the {\tt MAD} synthetic tasks, when using evaluation accuracy as a performance metric (left) or evaluation loss (right).}
    \label{appendix:fig:mad-scores-per-task}
\end{figure*}

\clearpage
\newpage

\subsubsection{Performance on Individual Tasks}

\begin{figure*}[!h]
    \centering
    \vspace{-0.1in}
    \includegraphics[width=0.75\textwidth]{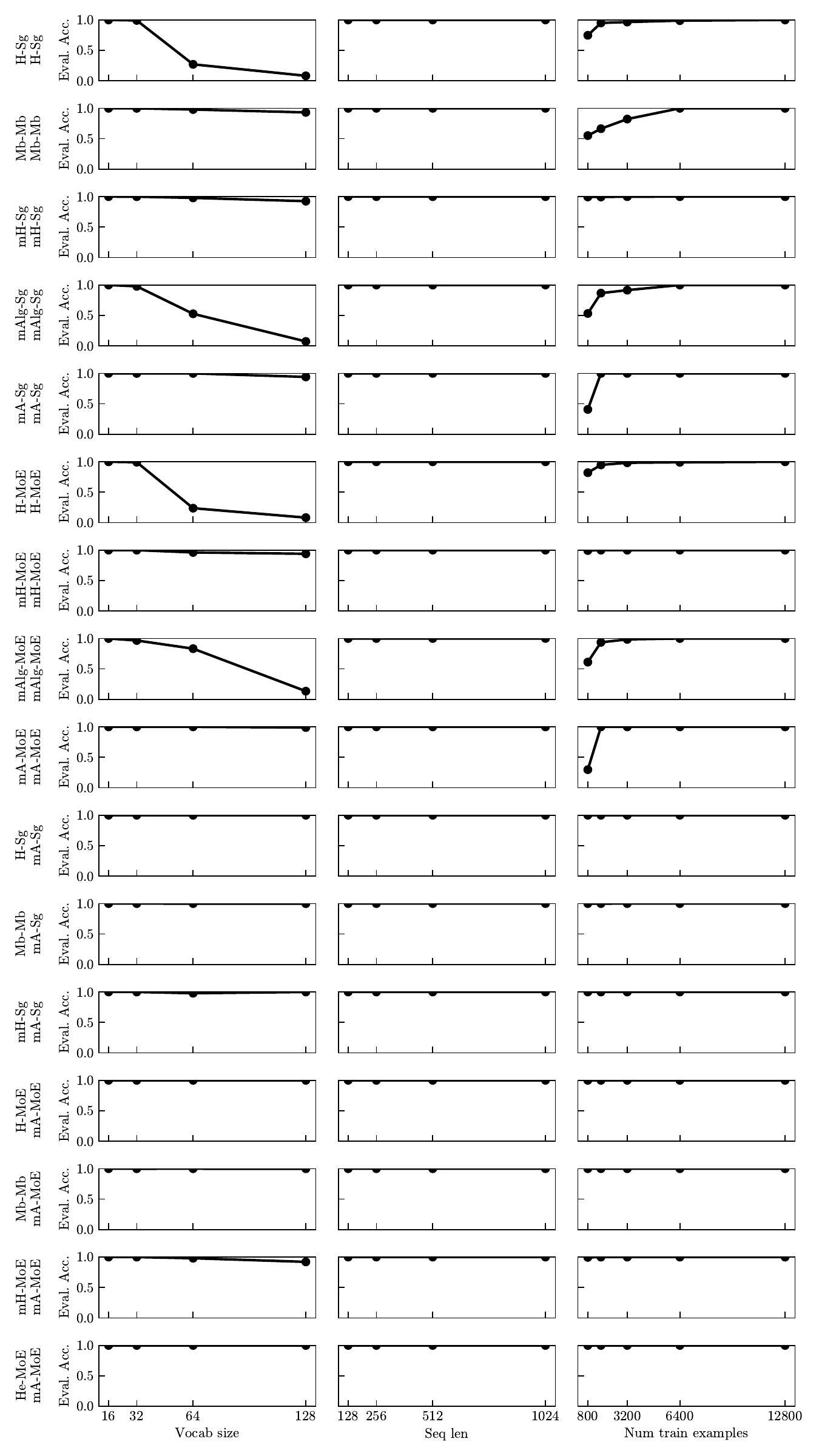}
    \vspace{-0.1in}
    \caption{\textbf{In-context recall task} model performances. {\tt H}: Hyena, {\tt Mb}: Mamba, {\tt Alg}: Gated Lin. Attention, {\tt A}: Attention, {\tt He}: Hyena Experts, {\tt Sg}: SwiGLU, {\tt MoE}: Mixture of Experts MLP, {\tt m\{H,A,Alg\}}: multi-headed model variants.}
    \label{appendix:fig:runs-context}
\end{figure*}

\clearpage
\newpage

\begin{figure*}[!h]
    \centering
    \vspace{-0.1in}
    \includegraphics[width=0.75\textwidth]{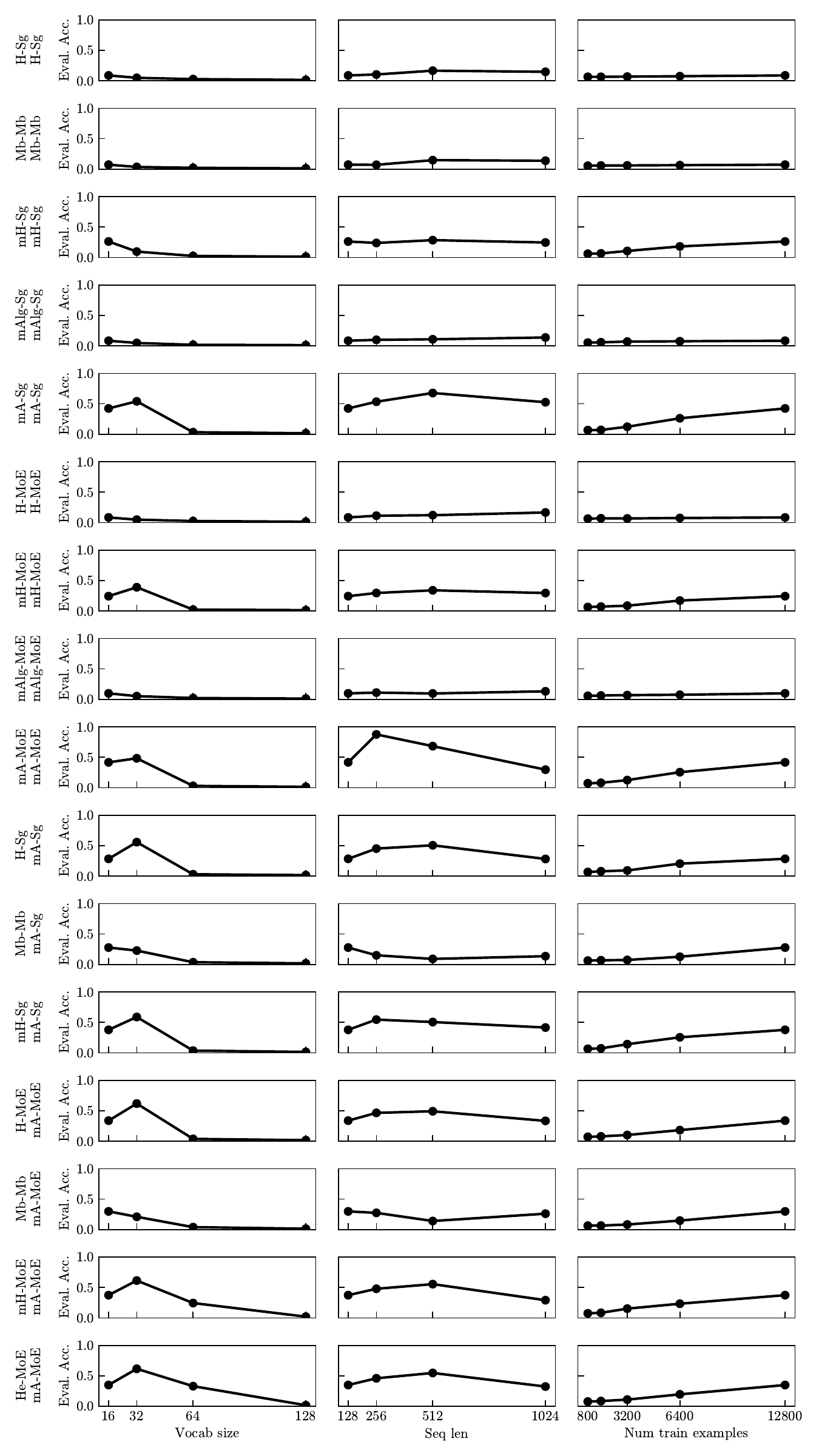}
    \vspace{-0.1in}
    \caption{\textbf{Fuzzy in-context recall task} model performances. {\tt H}: Hyena, {\tt Mb}: Mamba, {\tt Alg}: Gated Lin. Attention, {\tt A}: Attention, {\tt He}: Hyena Experts, {\tt Sg}: SwiGLU, {\tt MoE}: Mixture of Experts MLP, {\tt m\{H,A,Alg\}}: multi-headed model variants.}
     \label{appendix:fig:runs-fuzzy}
\end{figure*}

\clearpage
\newpage

\begin{figure*}[!h]
    \centering
    \vspace{-0.1in}
    \includegraphics[width=\textwidth]{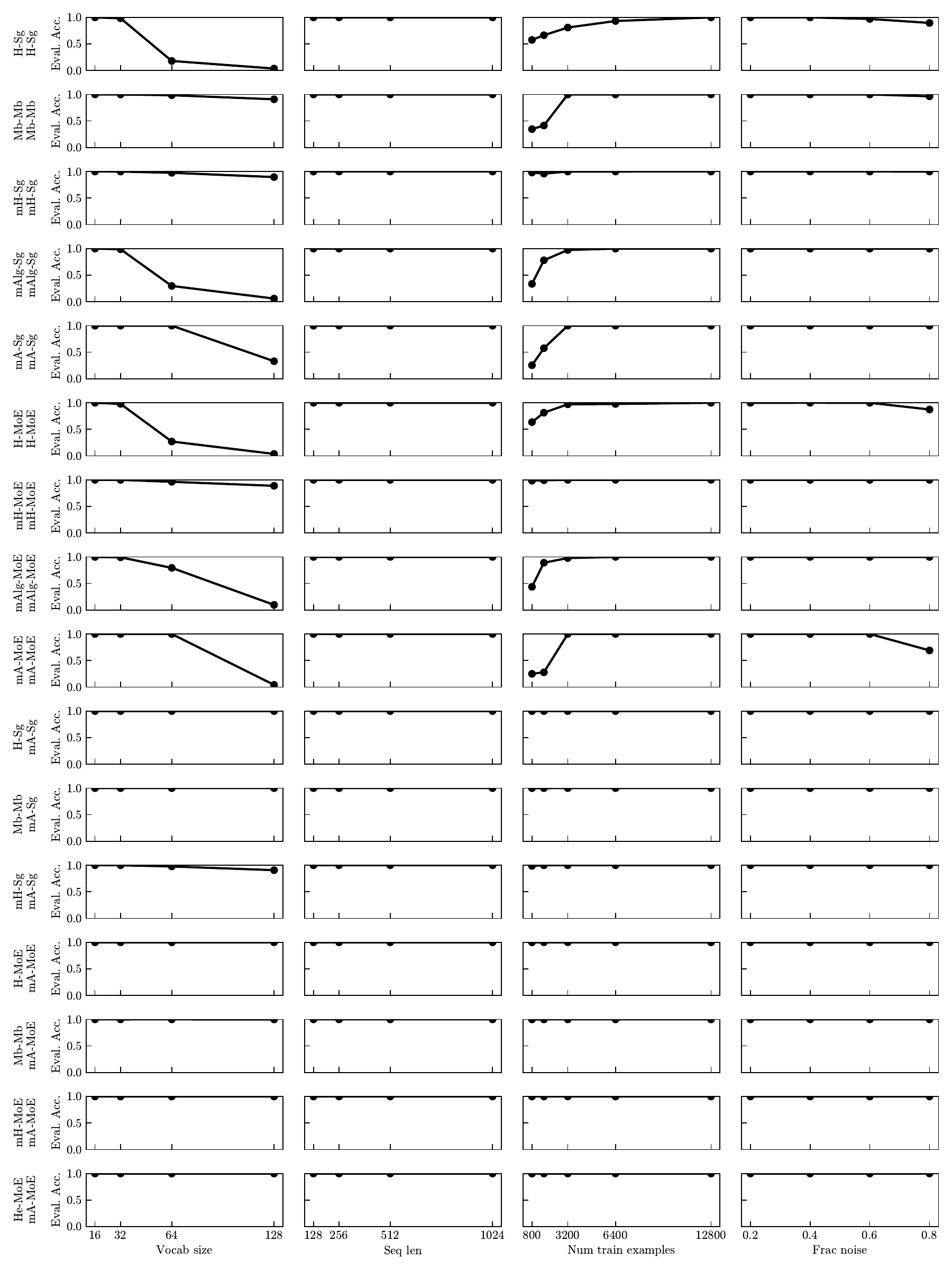}
    \vspace{-0.1in}
    \caption{\textbf{Noisy in-context recall task} model performances. {\tt H}: Hyena, {\tt Mb}: Mamba, {\tt Alg}: Gated Lin. Attention, {\tt A}: Attention, {\tt He}: Hyena Experts, {\tt Sg}: SwiGLU, {\tt MoE}: Mixture of Experts MLP, {\tt m\{H,A,Alg\}}: multi-headed model variants.}
    \label{appendix:fig:runs-noisy}
\end{figure*}

\clearpage
\newpage

\begin{figure*}[!h]
    \centering
    \vspace{-0.1in}
    \includegraphics[width=\textwidth]{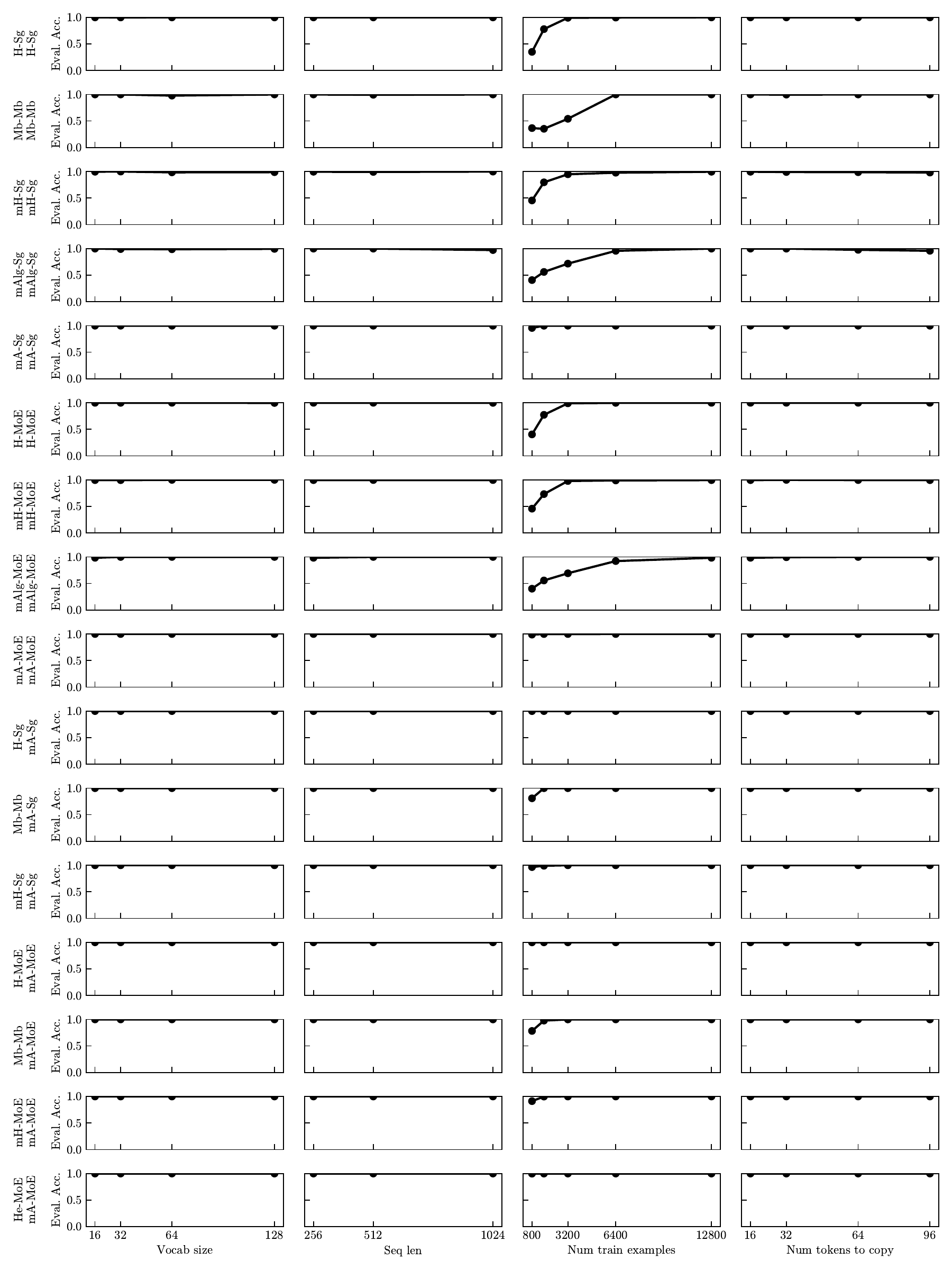}
    \vspace{-0.1in}
    \caption{\textbf{Selective Copying} model performances. {\tt H}: Hyena, {\tt Mb}: Mamba, {\tt Alg}: Gated Lin. Attention, {\tt A}: Attention, {\tt He}: Hyena Experts, {\tt Sg}: SwiGLU, {\tt MoE}: Mixture of Experts MLP, {\tt m\{H,A,Alg\}}: multi-headed model variants.}
    \label{appendix:fig:runs-copy}
\end{figure*}

\clearpage
\newpage

\begin{figure*}[!h]
    \centering
    \vspace{-0.1in}
    \includegraphics[width=0.75\textwidth]{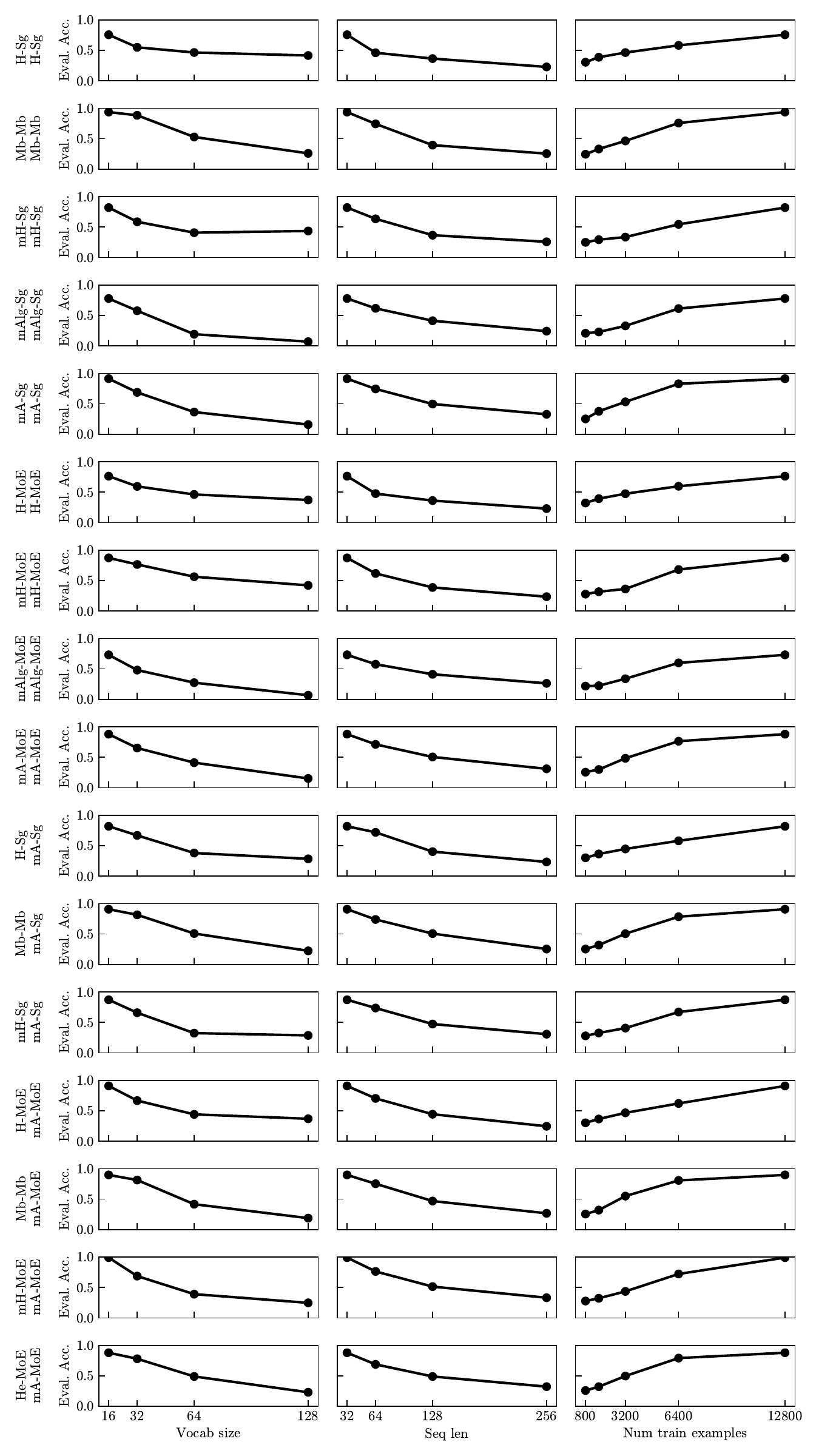}
    \vspace{-0.1in}
    \caption{\textbf{Compression} model performances. {\tt H}: Hyena, {\tt Mb}: Mamba, {\tt Alg}: Gated Lin. Attention, {\tt A}: Attention, {\tt He}: Hyena Experts, {\tt Sg}: SwiGLU, {\tt MoE}: Mixture of Experts MLP, {\tt m\{H,A,Alg\}}: multi-headed model variants.}
    \label{appendix:fig:runs-compression}
\end{figure*}

\clearpage
\newpage

\begin{figure*}[!h]
    \centering
    \vspace{-0.1in}
    \includegraphics[width=0.275\textwidth]{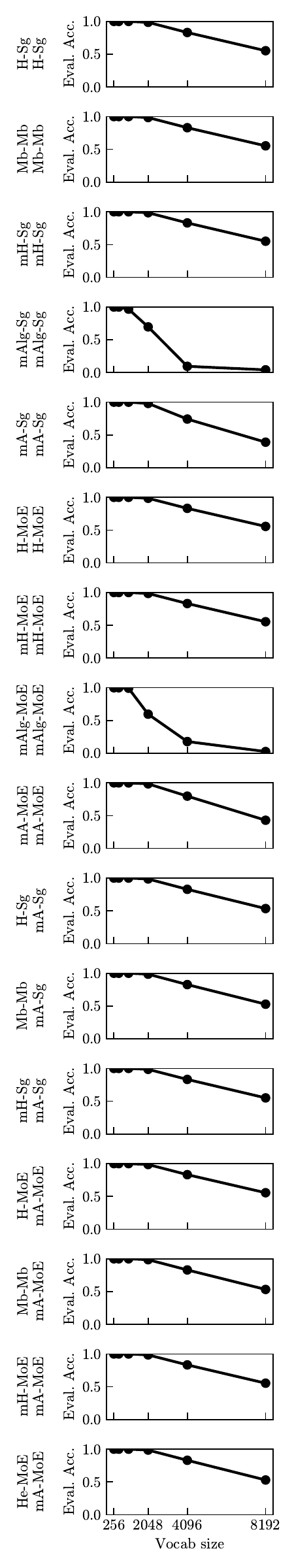}
    \vspace{-0.1in}
    \caption{\textbf{Memorization} model performances. {\tt H}: Hyena, {\tt Mb}: Mamba, {\tt Alg}: Gated Lin. Attention, {\tt A}: Attention, {\tt He}: Hyena Experts, {\tt Sg}: SwiGLU, {\tt MoE}: Mixture of Experts MLP, {\tt m\{H,A,Alg\}}: multi-headed model variants.}
    \label{appendix:fig:runs-memorize}
\end{figure*}

\clearpage
\newpage

\clearpage
\newpage

\begin{figure}
    \centering
    \vspace{-0.1in}
    \includegraphics[width=\columnwidth]{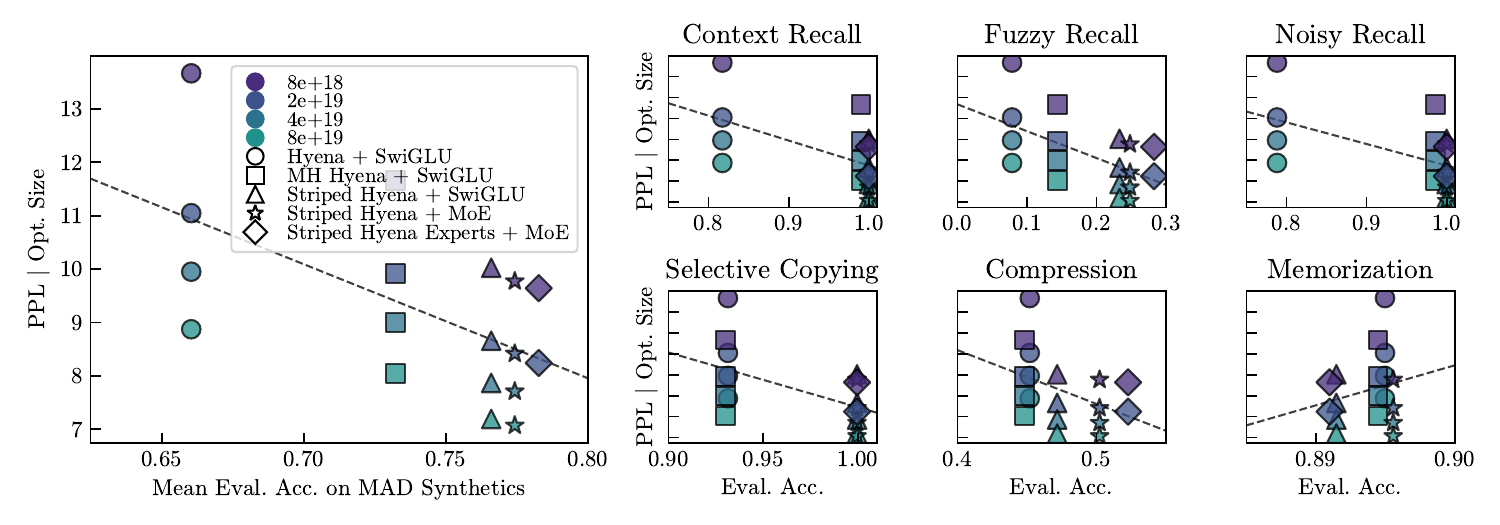}
    \vspace{-0.2in}
    \caption{Improved performance on MAD synthetics correlates with better compute-optimal perplexity on
    The Pile across IsoFLOP groups. We highlight progressively improved versions of Hyena that were designed with the MAD pipeline.}
    \vspace{-0.1in}
    \label{appendix:fig:mad-to-scale-hyena-full}
\end{figure}

\begin{figure}
    \centering
    \vspace{-0.1in}
    \includegraphics[width=\columnwidth]{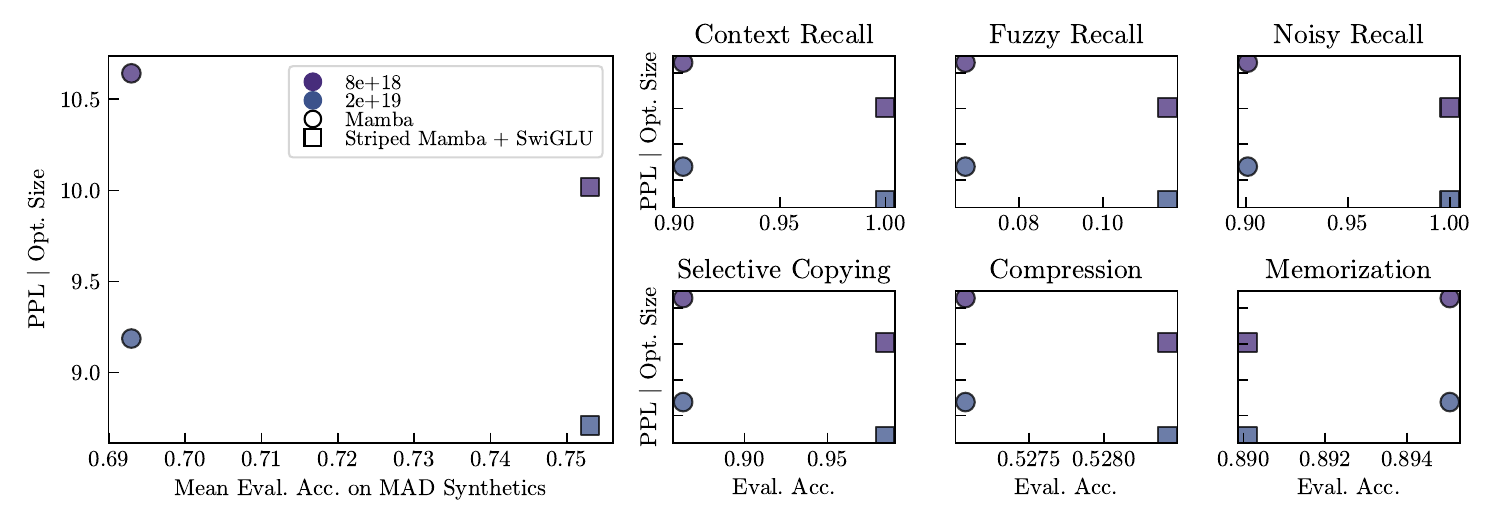}
    \vspace{-0.2in}
    \caption{Replication of Fig.~\ref{appendix:fig:mad-to-scale-hyena-full} for the Mamba and Striped Mamba architectures and IsoFLOP groups 8e18 and 2e19.}
    \vspace{-0.1in}
    \label{appendix:fig:mad-to-scale-mamba}
\end{figure}

\clearpage
\newpage
\section{Scaling Laws}\label{appendix:scaling-laws}

We design our model topologies starting from previous compute-optimal scaling results for Transformers ~\cite{sardana2023beyond}, and selecting the number of layers (depth) and width to cover a range of parameters from $8e6$ to $7e9$ parameters (see Table \ref{table:model_sizes}). The depth and width are generally fixed across models, which result in minor parameter count differences (except for the mixture of experts models where a distinction between total and active parameters must be made, see Tables \ref{table:moh_widths} and \ref{table:moe_widths}). To compare how each model scales, we control for several compute budgets (IsoFLOP groups): 4e18, 8e18, 2e19, 4e19, 8e19, 2e20, 5e20, 2e21. We linearly interpolate learning rates from common settings at $150e6, 350e6, 1.3e9, 3e9$ and $7e9$ model sizes, obtaining a linearly inverse relationship with model size. Batch size is scaled (increased) in discrete steps, with larger training FLOPs using larger batch sizes.

For state-optimal scaling results, we obtain the optimal model size from the compute-optimal frontier, then compute the dynamic and fixed state dimensions of the closest model size available in the set of results. 

\subsection{Training Details}

We control for key hyperparameters across all models, including batch size (Table \ref{table:batch_size}), learning rate (Table \ref{table:model_sizes}) and scheduler. Most models were trained on a single node. For larger IsoFLOP groups, we trained in a multinode distributed training with tensor parallelism. We used a cosine decay learning rate scheduler, with warm up using 1\% the number of training steps, and the minimum decay to reach 10\% of the max learning rate.

\begin{table}[ht]
    \centering
    \caption{Batch sizes by IsoFLOP group. For very small models ($<$54M) parameters, batch size 262k is used.}
    \begin{sc}
    \begin{tabular}{cc}
        \hline
        IsoFLOP & Batch Size \\
        \hline
        4.0E+18 & 524k \\
        8.0E+18 & 524k \\
        2.0E+19 & 524k \\
        4.0E+19 & 524k \\
        8.0E+19 & 524k \\
        2.0E+20 & 1M \\
        5.0E+20 & 1M \\
        2.0E+21 & 2M \\
    \hline
    \end{tabular}
    \end{sc}
    \label{table:batch_size}
\end{table}

\begin{figure}
    \centering
    \vspace{-0.2in}
    \includegraphics[width=0.45\linewidth]{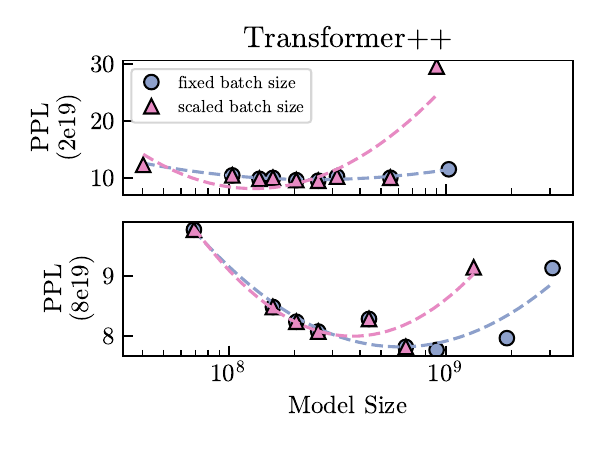}
    \vspace{-0.2in}
    \caption{Increasing batch size with compute FLOPS can shift the compute-efficient frontier. When increasing batch size after $10^9$ parameters (red), the IsoFLOP curve underestimates the performance of larger models, when compared to a fixed batch size (blue), shifting the optimum estimation towards smaller models.}
    \vspace{-0.1in}
    \label{fig:batch-size-effect}
\end{figure}

\subsection{Model architectures}\label{model-details}

We describe shared architecture details first, followed by model specific designs below. All models use a modern SwiGLU unit as the channel mixer, except for Mamba and StripedMamba (which merges the GLU block with the sequence mixer layer, resulting in twice the number of sequence mixers). We use RMSNorm \cite{zhang2019root} for normalization. All models tie the embedding layers. All sparsely activated layers use learned argmax routing.

\paragraph{Transformer{\tt ++}} Transformer{\tt ++} is state-of-the-art Transformer model, with rotary positional embeddings \cite{su2024roformer}, SwiGLU and RMSNorm.
    
\paragraph{Hyena} We use the original architecture \cite{poli2023hyena} with some improvements. The channel mixer is replaced with SwiGLU, we use RMSNorm, set weight decay to $0$ to all Hyena layer parameters. 
    
\paragraph{Multi-Head Hyena} We use a Hyena layer with heads as described by \cite{massaroli2023laughing}. We sweep across different head dimensions at the IsoFLOP group 2e19 to find an optimal head dimension (8), and use the same number for all other experiments.

\paragraph{StripedHyena} We use 3 striping schedule ratios: 1{\tt A}:1{\tt H}, 1{\tt A}:3{\tt H}, 1{\tt A}:11{\tt H}, where {\tt A}=Attention and {\tt H}=Hyena along model depth. In instances where the number of layers is not a multiple of the schedule, the ratio is repeated until the target depth is reached.
    
\paragraph{Mamba} Mamba doubles the number of sequence mixers, replacing the dedicated channel mixer, and uses a custom input-varying recurrence. Hyperparameters (state dimension $16$, expansion factor $2$, conv projection length $4$ and width of implicit network are sourced from the original implementation \cite{gu2023mamba})

\paragraph{StripedMamba} Similar to StripedHyena, we use the 3 striping ratio schedules to interleave attention at specified intervals along model depth.
    
\paragraph{StripedHyena-MoE} The StripedHyena-MoE replaces SwiGLU with a total of 8 experts and 2 active experts. We keep the same depth and model width in the mixer layer as baseline models, and adjust the MoE widths to match active parameters.
    
\paragraph{StripedHyena Experts-MoE} This model introduces expert in the Hyena sequence mixer at the output level, as described in the main text. We use a StripedHyena with striping ratio 1:11, and the following expert counts: total experts = 8, active experts = 2, total mixer experts = 8, active mixer experts = 2.


\subsection{Model sizes and training hyperparameters}

We show common model settings across all architectures by size in Table \ref{table:model_sizes}. We use Adam optimizer betas [0.9, 0.95], weight decay 0.1, and no dropout. All models are trained in mixed precision: {\tt bfloat16} with full precision for Hyena and Mamba convolution and recurrences.

\begin{table}[ht]
    \caption{Common settings across all architectures. For Mamba, we use the layer structure of {\tt Mb-Mb} following \cite{gu2023mamba}. Actual parameter counts vary slightly for each architecture..}
    \vspace{4mm}
    \centering 
    \begin{sc}
    \begin{tabular}{ccccccc}
    \hline
    Params (M) & d\_model & ffw\_size & kv\_size & n\_heads & n\_layers & learning rate \\ \hline
        8 & 128 & 336 & 64 & 2 & 4 & 9.77E-04 \\
        22 & 320 & 848 & 64 & 5 & 5 & 9.57E-04 \\
        38 & 448 & 1200 & 64 & 7 & 7 & 9.36E-04 \\
        54 & 512 & 1360 & 64 & 8 & 9 & 9.15E-04 \\
        70 & 576 & 1536 & 64 & 8 & 10 & 8.95E-04 \\
        102 & 640 & 1712 & 64 & 10 & 14 & 8.56E-04 \\
        118 & 704 & 1872 & 64 & 11 & 14 & 8.37E-04 \\
        134 & 768 & 2048 & 64 & 12 & 14 & 8.18E-04 \\
        150 & 768 & 2048 & 64 & 12 & 16 & 8.00E-04 \\
        163 & 768 & 2048 & 64 & 12 & 17 & 7.75E-04 \\
        175 & 768 & 2048 & 64 & 12 & 19 & 7.50E-04 \\
        196 & 832 & 2224 & 64 & 13 & 19 & 7.25E-04 \\
        217 & 832 & 2224 & 64 & 13 & 21 & 7.00E-04 \\
        251 & 896 & 2384 & 64 & 14 & 21 & 6.75E-04 \\
        278 & 896 & 2384 & 64 & 14 & 24 & 6.50E-04 \\
        306 & 960 & 2560 & 64 & 15 & 24 & 6.25E-04 \\
        350 & 1024 & 2736 & 64 & 16 & 24 & 6.00E-04 \\
        440 & 1152 & 3072 & 64 & 18 & 24 & 5.66E-04 \\
        536 & 1280 & 3408 & 64 & 20 & 24 & 5.33E-04 \\
        641 & 1408 & 3760 & 128 & 11 & 24 & 5.00E-04 \\
        756 & 1536 & 4096 & 128 & 12 & 24 & 4.75E-04 \\
        881 & 1664 & 4432 & 128 & 13 & 24 & 4.55E-04 \\
        1010 & 1792 & 4784 & 128 & 14 & 24 & 4.33E-04 \\
        1160 & 1920 & 5120 & 128 & 15 & 24 & 4.15E-04 \\
        1200 & 1920 & 5120 & 128 & 15 & 25 & 4.11E-04 \\
        1300 & 2048 & 5456 & 128 & 16 & 24 & 4.00E-04 \\
        1600 & 2176 & 5808 & 128 & 17 & 26 & 3.84E-04 \\
        1900 & 2304 & 6144 & 128 & 18 & 28 & 3.67E-04 \\
        2250 & 2432 & 6480 & 128 & 19 & 30 & 3.47E-04 \\
        2400 & 2560 & 6832 & 128 & 20 & 29 & 3.39E-04 \\
        2640 & 2560 & 6832 & 128 & 20 & 32 & 3.25E-04 \\
        3100 & 2688 & 7168 & 128 & 21 & 34 & 3.00E-04 \\
        4200 & 3072 & 8192 & 128 & 24 & 36 & 2.72E-04 \\
        5200 & 3328 & 8880 & 128 & 26 & 38 & 2.46E-04 \\
        7000 & 3712 & 9904 & 128 & 29 & 41 & 2.00E-04 \\
    \hline
    \end{tabular}
    \end{sc}
    \label{table:model_sizes}
\end{table}

\begin{table}[ht]
    \centering
    \caption{MoE model sizes for StripedHyena. All MoE models use 8 total experts and 2 active experts. Other model settings for corresponding \textbf{active parameter} counts follow Table \ref{table:model_sizes}, including d\_model, n\_heads, n\_layers, ffw\_size, kv\_size, and learning rate.}
    \vspace{4mm}
    \begin{sc}
    \begin{tabular}{cccc}
    \hline
    Total Params (M) & Active Params & MoE Width \\ \hline
       194 &        102 &        1728 \\
       228 &        118 &        1856 \\
       270 &        134 &        2048 \\
       303 &        150 &        2048 \\
       319 &        163 &        2048 \\
       352 &        175 &        2048 \\
       404 &        196 &        2176 \\
       452 &        217 &        2240 \\
       512 &        251 &        2368 \\
       580 &        278 &        2368 \\
       667 &        306 &        2560 \\
       761 &        350 &        2752 \\
       950 &        440 &        2752 \\
       1160 &       536 &        3392 \\
       1390 &       641 &        3712 \\
       1660 &       756 &        4096 \\
       1940 &       881 &        4416 \\
       2230 &       1010 &       4736 \\
       2550 &       1160 &       5056 \\
       2910 &       1300 &       5440 \\
    \hline
    \end{tabular}
    \end{sc}

    \label{table:moe_widths}
\end{table}

\begin{table}[ht]
    \centering
    \caption{StripedHyena Expert model sizes, which all use 8 total experts and 2 active experts for both sequence mixing and GLU experts. Other model settings for corresponding \textbf{active parameter} counts follow Table \ref{table:model_sizes}, including d\_model, n\_heads, n\_layers, ffw\_size, kv\_size, and learning rate.}
    \begin{sc}
    \vspace{4mm}
    \begin{tabular}{ccccc}
    \hline
    Total Params (M) & Active Params & Expert Width & Expert Total Width & MoE Width \\ \hline
       241 &        101 &        80 &       640 &        2368 \\
       290 &        119 &        88 &       704 &        2624 \\
       337 &        137 &        96 &       768 &        2816 \\
       386 &        153 &        96 &       768 &        2880 \\
       408 &        160 &        96 &       768 &        2880 \\
       452 &        174 &        96 &       768 &        2880 \\
       520 &        199 &        104 &      832 &        3072 \\
       570 &        215 &        104 &      832 &        3072 \\
       661 &        248 &        112 &      896 &        3328 \\
       749 &        277 &        112 &      896 &        3328 \\
       860 &        315 &        120 &      960 &        3584 \\
       965 &        352 &        128 &      1024 &        3776 \\
       1220 &        441 &        144 &      1152 &        4288 \\
       1500 &        535 &        160 &      1280 &        4736 \\
       1810 &        641 &        176 &      1408 &        5216 \\
       2140 &        757 &        192 &      1536 &        5696 \\
       2510 &        882 &        208 &      1664 &        6176 \\
       2880 &        1010 &       224 &      1792 &        6592 \\
       3320 &        1160 &       240 &       1920 &        7104 \\
       3790 &        1310 &       256 &      2048 &        7616 \\
    \hline
    \end{tabular}
    \end{sc}
    \label{table:moh_widths}
\end{table}

\subsection{FLOP calculation}
\label{sec:flop}

We provide FLOP calculators for each model architecture explored in this study. Notation is provided in \ref{table:notation_flop}.

\begin{table}[ht]
    \centering
    \caption{Notation for FLOP calculation.}
    \vspace{2mm}
    \begin{tabular}{cc}
    \toprule
    Notation & Description \\ \toprule
    $C$ & Model FLOP cost per token   \\  
    $N$ & Number of layers \\
    $L$ & Sequence length \\
    $D$ & Model width \\ 
    $V$ & Vocabulary size \\ 
    $H$ & Number of heads \\ 
    $D_{\tt glu}$ & Width of GLU (reverse bottleneck) \\
    $D_{\tt moe}$ & Width of MoE expert \\
    $D_{\tt dt}$ & Width of bottleneck in Mamba featurization \\
    $D_{\tt moh}$ & Width of Hyena expert \\
    $D_{\tt glu}$ & Width of GLU (reverse bottleneck) \\
    $A_{\tt moe}$ & Number of MoE experts \\
    $A_{\tt moh}$ & Number of Hyena experts \\
    $G_{\tt moe}$ & Number of active MoE experts \\
    $G_{\tt moh}$ & Number of active Hyena experts \\
    $S_{\tt hyena}$ & filter order \\
    $S_{\tt mamba}$ & state dimension \\
    $E$ & projection expansion factor \\
    \bottomrule
    \end{tabular}
    \label{table:notation_flop}
\end{table}

\subsubsection{Transformer{\tt ++}}

\begin{itemize}
    \item Embedding layers: $4 L D V$
    \item MHA
        \begin{itemize}
            \item \textbf{projections}: $6 L D^2$
            \item \textbf{attention}: $4 L^2 D + 2 H L^2$
            \item \textbf{out layer}: $2 L D^2$
        \end{itemize}
    \item GLU
        \begin{itemize}
            \item $6 L D D_{\tt glu}$ 
        \end{itemize}    
\end{itemize}

\subsubsection{Hyena}
GLU and embedding calculation is the same as Transformer{\tt ++}.

\begin{itemize}
    \item Sequence Mixer
        \begin{itemize}
            \item \textbf{projections}: $6 L D^2$
            \item \textbf{convs on projections}: $18 L D$
            \item \textbf{featurization}: $S_{\tt hyena} L D$\footnote{Other filter parametrizations e.g., canonical via rational functions, scale with the order $\mathcal{O}(S_{\tt hyena} D L \log_2(L))$.}
            \item \textbf{convolution and gates}: $10 L \log_2(L) D + 4 L D$ 
            \item \textbf{out layer}: $2 L D^2$          
        \end{itemize}
\end{itemize}

\subsubsection{Multi-Head Hyena}

\begin{itemize}
    \item Sequence Mixer
        \begin{itemize}
            \item \textbf{projections}: $6 L D^2$
            \item \textbf{convs on projections}: $18 L D$
            \item \textbf{featurization}: $S_{\tt hyena} L H$
            \item \textbf{convolution and gates}: $10 L \log_2(L) D^2 / H + 4 L D^2 / H$  
            \item \textbf{out layer}: $2 L D^2$          
        \end{itemize}
\end{itemize}

\subsubsection{StripedHyena}

FLOPS of StripedHyena are determined by summing the FLOPS of Hyena-GLU and MHA-GLU, with the mixing ratios specified by the particular instance of the model.

\subsubsection{Mamba}

\begin{itemize}

    \item Sequence Mixer
        \begin{itemize}
            \item \textbf{projections}: $4 L D^2 E$
            \item \textbf{short conv}: $6 L D E$
            \item \textbf{featurization}: $2 L D E (D_{\tt dt} + 2 S_{\tt mamba}) + 2 L D E D_{\tt dt}$
            \item \textbf{associative scan}: $2 L D E S_{\tt mamba}$\footnote{Estimate assumes "most efficient" scan algorithm in terms of FLOPS (but not latency). In practice, the constant may be larger.}
            \item \textbf{out layer}: $2 L D^2 E$
        \end{itemize}
    \item No separate GLU block (2x the sequence mixers).

\end{itemize}

\subsubsection{StripedMamba}

FLOPS of StripedMamba are determined by summing the FLOPS of Mamba-Mamba and MHA-GLU, with the mixing ratios specified by the particular instance of the model.

\subsubsection{StripedHyena-MoE}

\begin{itemize}

    \item Sequence mixer
        \begin{itemize}
            \item Same as StripedHyena
        \end{itemize}
    \item SwiGLU MoE (replaces MLP block)
        \begin{itemize}
            \item \textbf{router}: $L D A_{\tt moe}$
            \item \textbf{up projections} $4 D D_{\tt moe} A_{\tt moe}$
            \item \textbf{down projection (sparse)} $2 D D_{\tt moe} G_{\tt moe}$
        \end{itemize}
\end{itemize}

\subsubsection{StripedHyena Experts + MoE}

Model has experts in both sequence mixers (Hyena) and GLU layers. In attention layers, Transformer{\tt ++} sequence mixer (MHA) FLOPS are used. The idea of Hyena experts is to select via a router (softmax - argmax selection) $G_{\tt moh}$ smaller Hyena experts, and run computation only on those dimensions. Equivalently, this can be seen as adaptively choosing a subset of states, using the input sequence.

\begin{itemize}
    \item Hyena experts
        \begin{itemize}
            \item \textbf{router}: $L D A_{\tt moh}$
            \item \textbf{projections}: $6 L D^2$
            \item \textbf{convs on projections}: $18 L D$
            \item \textbf{featurization}: $S_{\tt hyena} L D_{\tt moh} G_{\tt moh}$
            \item \textbf{convolution and gates}: $10 L \log_2(L) D_{\tt moh} G_{\tt moh} + 4 L D_{\tt moh} G_{\tt moh}$ 
            \item \textbf{out layer}: $2 L D_{\tt moh} D$          
        \end{itemize}
\end{itemize}

\section{Extended Scaling Results}

\subsection{Optimal hybridization topologies}
We observe the topology of hybrid architectures to have significant effect on their downstream performance. In MAD tests, interleaving schedules for StripedHyena, with gated convolution followed by attention, outperform schedules attention followed by gated convolution.

Table \ref{table:striping-topology} provides ablations on the perplexity at larger scales. A variety of topologies achieve best perplexity, including chunked interleaving (6H:6A) and an \textit{encoder-decoder} topology (6H:12A:6H), where Hyena layers surround a block of attention layers.

For all other experiments in the paper, including scaling laws, we adopt a simple 1H:1A topology for simplicity, as that is already seen to outperform other architectures in compute-optimal and state-optimal scaling.

\begin{table}[!h]
\caption{Topology ablation for StripedHyena (750M at 2e19 FLOPS on The Pile). {\tt H} and {\tt A} indicate Hyena and MHA layers, respectively.}
\label{table:striping-topology}
\vskip 0.15in
\begin{center}
\begin{tabular}{lcccr}
\toprule
{\sc Topology} & {\sc Perplexity}\\
\midrule
(1H:1A) $\times 12$ & $9.52$ \\
(2H:2A) $\times 6$ & $9.32$ \\
(3H:3A) $\times 4$ & $9.33$ \\
(4H:4A) $\times 3$ & $9.37$ \\
(6H:6A) $\times 2$ & $9.28$ \\
12H:12A & 9.41 \\
\midrule
(2H:4A:4H:2A) $\times 2$ & $9.25$ \\
(H:5A:5H:A) $\times 2$ & $9.31$ \\
4H:10A:8H:2A &$9.33$ \\
4H:12A:8H & $9.31$ \\
6H:12A:6H & $9.30$ & \\
8H:12A:4A & $9.35$ \\
\bottomrule
\end{tabular}
\end{center}
\vskip -0.1in
\end{table}

\subsection{Byte-level scaling laws}\label{appendix:byte}

\begin{figure*}[!h]
    \centering
    \includegraphics[width=\textwidth]{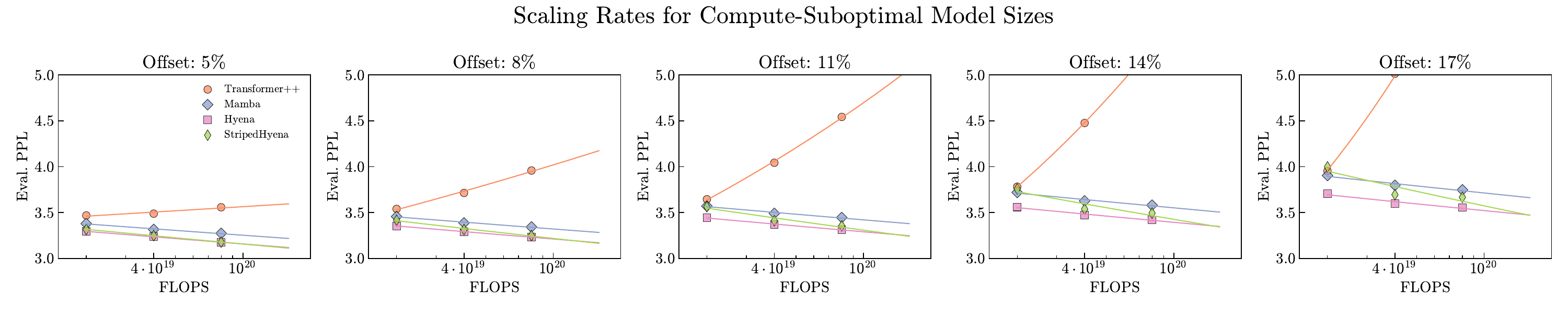}
    \vspace{-4mm}
    \caption{Scaling off the compute-optimal frontier on DNA data. We verify the perplexity scaling at model sizes with a percentage offset from the optimal model size at each FLOP budget. In particular, we train a \% offset smaller model, for longer. Transformers do not scale well to the overtraining regime.}
    \label{fig:dna-off-optimal}
\end{figure*}

We report additional results for scaling laws on DNA sequences. We trained all models on $8$k sequence length, using model hyperparameters detailed in \ref{model-details}. The model rankings are different from subword tokenized language data. We also compare architecture performance outside the compute-optimal frontier, namely with allocations of the computational budget are suboptimal but common in practice, such as overtraining smaller models \ref{fig:dna-off-optimal}. 

\begin{figure*}[!h]
    \centering
    \includegraphics[width=\textwidth]{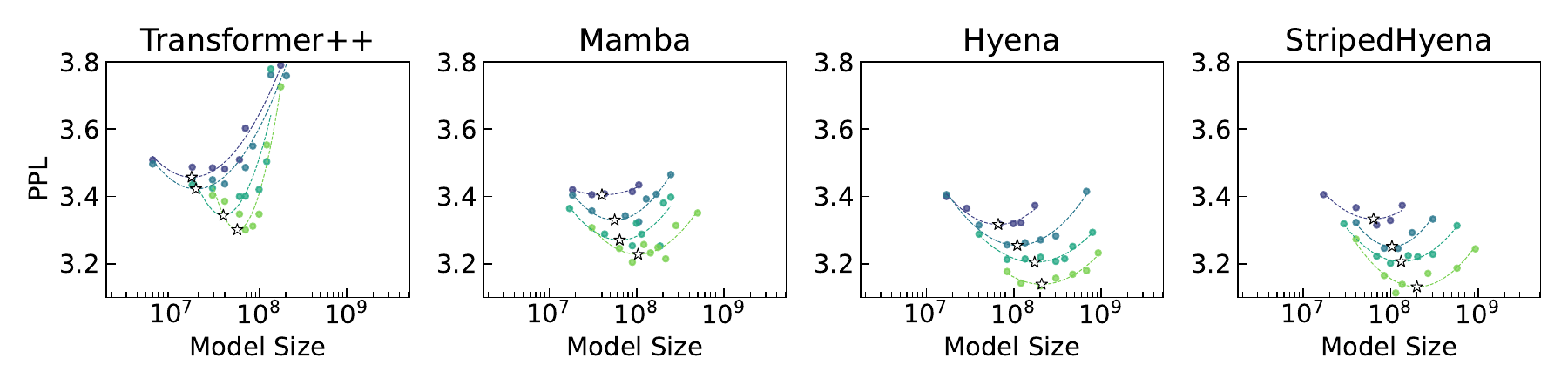}
    \vspace{-8mm}
    \caption{Pretraining compute-optimal scaling on DNA sequences, with byte-level tokenization (nucleotide resolution).}
    \label{fig:dna-isoflop}
\end{figure*}

\end{document}